\documentclass{article}

\usepackage{microtype}
\usepackage{graphicx}
\usepackage[format=hang]{caption}
\usepackage{subcaption}
\usepackage{booktabs} 
\usepackage{amsmath}
\usepackage{amssymb}
\usepackage{bm}
\usepackage{gensymb}
\usepackage{csquotes}
\usepackage{centernot}
\usepackage{nicefrac}
\usepackage{textcomp}
\usepackage{multicol}
\usepackage{float}
\usepackage{siunitx}
\usepackage{soul}
\usepackage{graphicx}
\usepackage{grffile}
\usepackage{multirow}
\usepackage{url}

\usepackage{hyperref}

\usepackage[accepted]{icml2021}

\newcommand{\twodots}{\mathinner {\ldotp \ldotp}}

\icmltitlerunning{Attend2Pack}

\begin{document}

\twocolumn[
\icmltitle{
   Attend2Pack: \\
   Bin Packing through Deep Reinforcement Learning with Attention
}



\icmlsetsymbol{equal}{*}

\begin{icmlauthorlist}
\icmlauthor{Jingwei Zhang}{equal,dorabot}
\icmlauthor{Bin Zi}{equal,dorabot}
\icmlauthor{Xiaoyu Ge}{dorabot,anu}
\end{icmlauthorlist}

\icmlaffiliation{dorabot}{Dorabot Inc, Shenzhen, Guangdong, China}
\icmlaffiliation{anu}{Australian National University, Canberra, Australia}

\icmlcorrespondingauthor{}{jingwei.zhang, bin.zi@dorabot.com}

\icmlkeywords{Machine Learning, ICML}

\vskip 0.3in
]



\printAffiliationsAndNotice{\icmlEqualContribution} 

\begin{abstract}

This paper seeks to tackle the 
bin packing problem (BPP)
through a learning perspective.
Building on self-attention-based encoding
and deep reinforcement learning algorithms,
we propose a new end-to-end
learning model for this task of interest.
By decomposing the combinatorial action space,
as well as 
utilizing a new training technique 
denoted as prioritized oversampling,
which is a general scheme to speed up on-policy learning,
we achieve state-of-the-art performance 
in a range of
experimental settings.
Moreover,
although the proposed approach
$\mathrm{attend2pack}$
targets offline-BPP,
we strip 
our method 
down to 
the strict online-BPP setting
where
it is also able to achieve 
state-of-the-art performance.
With a set of ablation studies 
as well as 
comparisons against a range of previous works,
we hope to offer as 
a valid baseline approach 
to this field of study.

\end{abstract}

\section{Introduction}
\label{sec:intro}

\subsection{Background}

At loading docks,
the critical problem faced by the workers everyday
is how to 
minimize the number of containers needed 
to pack all cargos
in order to reduce transportation expenses;
at factories and warehouses, 
palletizing robots
are programmed to stack parcels 
onto pallets 
for efficient shipping.
The central problem in these 
real life scenes is the 
bin packing problem (BPP), 
which is having an evergrowing impact
on logistics, trading and economy 
under globalization and 
the boosting of e-commerce. 

Given a set of items 
and a number of bins, 
BPP
seeks for the packing configuration
under which the number of bins needed is minimized 
while utilization 
(Sec.\ref{sec:methods})
is maximized.

There are in general two types of bin packing problems, 
\textit{online}-BPP and \textit{offline}-BPP,
depending on whether
the upcoming loading items are known in advance. 
Specifically, 
in offline-BPP, 
the information of all items to be packed
are known beforehand,
while items arrive 
one by one in online-BPP
and the agent is only aware of the current item at hand.
The typical application
of online-BPP
is palletizing,
e.g.,
instantaneously packing 
items transported by a conveyor belt
onto a pallet with a fixed bottom size
\cite{zhao2020online}, 
where the number of items to be packed onto one pallet
is usually in the order of tens;
while the practical use cases of offline-BPP
include loading 
cargos into containers so as to be transported 
by trucks or freighters,
where the number of cargos per container could be
in the order of hundreds.
In this paper, 
we mainly focus on offline-BPP,
especially the single container loading problem (SCLP)
\cite{gehring2002parallel,huang2009caving,zhu2012six}
in which the goal is to pack 
all items into a bin
in such a way that utilization is maximized, 
while our proposed method
can be ablated to strict online-BPP settings
(Sec.\ref{sec:random-online-ablation}).

\subsection{Traditional Algorithms}

As a family of NP-hard problems
\cite{lewis1983computers},
the optimal solution of BPP 
cannot be found in polynomial time 
\cite{sweeney1992cutting,martello2000three,crainic2008extreme}.
Approximate methods 
and domain-specific heuristics
thus have been proposed 
to find near-optimal solutions.
Such methods include:
branch and bound 
\cite{martello1990knapsack,lysgaard2004new,lewis2008linear};
integer linear programming
\cite{dyckhoff1981new,cambazard2010propagating,salem2020new};
extreme-point-based constructive heuristics
\cite{crainic2008extreme}
where the corner of new items must be 
placed in contact with one of the corner points inside the container;
block-building-based heuristic
\cite{fanslau2010tree} 
that brings substantial performance gains,
in which items are first constructed into 
common building blocks then stacked into containers; 
as well as meta-heuristic algorithms
\cite{abdel2018improved,loh2008solving}.
However,
the general applicability of these algorithms
has been limited
either by high computational costs
or the requirement for domain experts 
to adjust the algorithm specifications 
for each newly encountered scenario.

\subsection{Learning-based Algorithms}

Inspired by the success brought by 
powerful deep function approximators
\cite{mnih2015human,he2016deep,vaswani2017attention},
BPP research has witnessed 
a surge of algorithms 
that approach this problem
from a learning perspective. 
As a combinatorial optimization problem,
although it requires exponential time 
to find the optimal solution 
in the 
combinatorial 
solution space,
it is relatively cheap 
to evaluate the quality of a given solution for BPP.
This makes the family of
policy gradient algorithms 
\cite{williams1992simple,sutton1999policy}
a good fit to such problems,
as given an evaluation signal
that is not necessarily differentiable,
policy gradient offers 
a principled way to guide search with gradients,
in contrast to gradient-free methods.
Another advantage of learning-based approaches
is that they are usually able to scale linearly 
to the input size,
and have satisfactory generalization capabilities.

Such attempts of utilizing 
deep reinforcement learning techniques 
for combinatorial optimization
starts from 
\cite{bello2016neural}
which tackles the traveling salesman problem (TSP)
with pointer networks
\cite{vinyals2015pointer}.
\cite{kool2018attention}
proposes an attention-based 
\cite{vaswani2017attention}
model that brings substantial improvements
in the vehicle routing problem (VRP) domain.


As for BPP,
\cite{zhao2020online,young2020smart,tanaka2020simultaneous}
focus on online-BPP and learn how to place each given box,
\cite{hu2017solving,duan2019multi}
learn to generate sequence order for boxes
while placing them using a fixed heuristic;
therefore these works 
do not target the full problem of offline-BPP.
Ranked reward (RR)
\cite{laterre2018ranked}
learns on the full problem
by conducting self-play under 
the rewards evaluated by ranking.
Although RR outperforms 
MCTS
\cite{browne2012survey}
and
the GUROBI solver
\cite{gurobi2018gurobi}
especially with large problem sizes,
it directly learns on the 
full combinatorial action space
which could pose major challenges 
for efficient learning.
\cite{zhao2020online,jiang2021solving}
also propose end-to-end approaches for offline-BPP,
however,
their state representation for encoding  
might not be optimized for BPP.
In this work,
we propose a new attention-based model
specifically designed for BPP.
To facilitate efficient learning, 
we decompose the combinatorial action space
and propose prioritized oversampling
to boost on-policy learning.
We conduct a set of ablation studies
and show that our method 
achieves state-of-the-art results 
in a range of comparisons.
\section{Methods}
\label{sec:methods}

We present our approach in the context of 3D bin packing
with single bin, 
from which its application to the 2D situation should be straightforward.

We formulate the bin packing problem (BPP) as a Markov decision process.
At the beginning of each episode,
the agent is given a set of $N$ boxes
each with dimensions 
($l^n$, $w^n$, $h^n$), 
as well as a bin of width $W$ and height $H$ 
with a length of 
$L=\sum_{n=1}^N\max(l^n, w^n, h^n)$
such is long enough to 
contain all these $N$ boxes 
under all variations of packing configurations,
as long as each box is in contact with 
at least one other box.
These together make up the input set of a 3D bin packing problem 
$
\mathcal{I}
=
\{
(l^1, w^1, h^1),
(l^2, w^2, h^2),
\ldots,
(l^{N}, w^{N}, h^{N}),
(L,
W,
H)
\}$.
The goal for the agent is to pack all $N$ boxes
into the bin in such a way
that the utility $r_{\mathrm{u}}\in[0,1]$ is maximized,
under geometry constraints
that no overlap between items is allowed.
The utility is calculated as
\begin{align}
    r_{\mathrm{u}}(\mathcal{C}_{\pi})
&=
    \frac{\sum_{n=1}^{N} l^n w^n h^n }{L_{\mathcal{C}_\pi} W H },
\end{align}
where 
$\mathcal{C}_{\pi}$
is a packing configuration 
(discussed in more details in Sec.\ref{sec:decomposing})
generated by following packing policy 
$\pi$,
and
$L_{\mathcal{C}_\pi}$ 
represents for this packing configuration
$\mathcal{C}_{\pi}$
the length of the minimum bounding box 
containing all the $N$ packed boxes inside of the bin.
The utility is given to the agent as a terminal reward $r_{\mathrm{u}}$
only upon when the last box has been packed into the bin,
meaning that the agent would receive no step or intermediate reward
during an episode.
We choose this terminal reward setup 
since it is non-trivial to design local step reward functions that 
would perfectly align with the global final evaluation signal. 
We also designed and experimented with other terminal reward formulations
but found utility to be a simple yet effective solution,
as well as the fact that it has a fixed range
which makes it easier to search for hyperparameters
that can work robustly across various datasets. 

When placing a box into the bin,
the box can be rotated $90^{\circ}$ along each of its three axes,
leading to a total number of $O=6$ possible orientations.
When it comes to the coordinate for placement inside a bin of
$L\times W\times H$,
there are also in total 
a maximum number of
$L\times W\times H$
possible locations 
to place a box
(we denote the rear-left-bottom corner of the bin as the origin
$(0,0,0)$
and define the location of a box $n$ inside the bin 
also by the coordinate 
$(x^n,y^n,z^n)$
of its rear-left-bottom corner).
This leads to a full action space of size
$N\times O\times L\times W\times H$
for the full bin packing problem.

\subsection{Decomposing the Action Space}
\label{sec:decomposing}
To deal with the potentially very large combinatorial action space,
we decompose the bin packing policy $\pi$ of the agent into two:
a sequence policy $\pi^{\mathrm{s}}$
that selects at each time step $t$
the next box to pack 
$s_t\in\{1,\ldots,N\}\setminus{s_{1:t-1}}$
(with action space 
$\mathcal{S}$ of size 
$\lvert\mathcal{S}\rvert=N$)
and a placement policy $\pi^{\mathrm{p}}$ 
that picks the orientation
and the location 
for the currently selected box
$p_t=(o^{s_t}, x^{s_t}, y^{s_t}, z^{s_t})$
(with action space 
$\mathcal{P}$
of size 
$\lvert\mathcal{P}\rvert=O\times L\times W\times H$).
We note that once a location $(x^{s_t}, y^{s_t})$ is selected for the current box,
its coordinate along the height dimension 
$z^{s_t}$
is also fixed because all items 
must have support along the height dimension;
also since we expect the packing to be compact as possible that
each box must be in contact with at least one other box,
then once
its coordinate 
$y^{s_t}$
along the width dimension is selected,
its coordinate 
$x^{s_t}$
along the length dimension
can be automatically located
as the rearest feasible point 
where the current box would be 
in contact with at least one other box or the rear wall of the bin,
and at the same time does not overlap with any other boxes.
Therefore,
the actual working action space for the placement policy 
$\pi^{\mathrm{p}}$ is of size
$\lvert\mathcal{P}\rvert=O\times W$,
that by selecting a placing coordinate
$y^{s_t}$
along the width dimension,
the other two coordinates for 
$s_t$
are located as the 
rearest-lowest
feasible point.

This decomposition can be regarded as casting 
the full joint combinatorial action space among two agents,
formulating BPP as a fully-cooperative 
multi-agent reinforcement learning task
\cite{panait2005cooperative,busoniu2008comprehensive}.
We note that now the learning system could be 
more properly modeled as a sequential Markov game
\cite{littman1994markov,lanctot2017unified,yang2020sequential} 
with two agents:
a sequence agent with action space 
$\mathcal{S}$
and a placement agent with action space
$\mathcal{P}$
sharing the same reward function.




Following this decomposition,
the joint policy $\pi$ 
given input $\mathcal{I}$
is factorized as 
\begin{align}
    \pi
&
        \big(
            s_1, p_1, s_2, p_2, \ldots, s_N, p_N 
            \hspace{1pt}\big\vert\hspace{1pt} 
            \mathcal{I}
        \big)
=\notag\\&
    \prod_{t=1}^N
    \pi^{\mathrm{s}}
        \Big(
            s_t \Big\vert f^{\mathrm{s}}_{\mathcal{I}}(s_{1:t-1}, p_{1:t-1})
        \Big)
    \pi^{\mathrm{p}}
        \Big(
            p_t \Big\vert f^{\mathrm{p}}_{\mathcal{I}}(s_{1:t}, p_{1:t-1})
        \Big),
\label{eq:pi}
\end{align}
in which the sequence policy $\pi^{\mathrm{s}}$
and the placement policy $\pi^{\mathrm{p}}$
together generate the full solution configuration
$\mathcal{C}_{\pi^{\mathrm{s}},\pi^{\mathrm{p}}}
=
\{s_1, p_1, s_2, p_2, \ldots, s_N, p_N\}$
in an interleaving manner;
$f_{\mathcal{I}}$ is the encoding function that maps 
(partially completed) configurations 
into state representations
for a particular input set $\mathcal{I}$,
which will be discussed below.

We will denote the deep network parameters for encoding as 
$\bm{\theta}^\mathrm{e}$
(Sec.\ref{sec:encoding}),
the parameters for decoding the sequence policy as
$\bm{\theta}^\mathrm{s}$
(Sec.\ref{sec:decoding-sequence}),
and 
those for the placement policy as
$\bm{\theta}^\mathrm{p}$
(Sec.\ref{sec:decoding-placement});
their aggregation 
is denoted as
$\bm{\theta}$.
Dependencies on these parameters 
are sometimes omitted in notation.



\subsection{Encoding}
\label{sec:encoding}

\subsubsection{Box Embeddings}
At the beginning of each episode,
the dimensions of each box in the input
are first encoded through a linear layer 
\begin{align}
    \bar{\bm{\mathsf{b}}}^n 
= 
    \mathrm{Linear}
    (l^n,w^n,h^n).
\end{align}
The set of $N$ such embeddings 
$\bar{\bm{\mathsf{b}}}$
($\lvert \bar{\bm{\mathsf{b}}} \rvert=d$)
are then passed through several 
multi-head 
(with $M$ heads)
self-attention layers
\cite{vaswani2017attention,radford2019language},
with each layer containing the following operations 
\begin{align}
    \tilde{\bm{\mathsf{b}}}^n
&=
    \bar{\bm{\mathsf{b}}}^n
+ 
    \mathrm{MHA}\big(
    \mathrm{LN}\big(
            \bar{\bm{\mathsf{b}}}^1,
            \bar{\bm{\mathsf{b}}}^2,
            \ldots,
            \bar{\bm{\mathsf{b}}}^N
    \big)
    \big),
\label{eq:att-att}
\\
    \bm{\mathsf{b}}^n
&=
    \tilde{\bm{\mathsf{b}}}^n
+ 
    \mathrm{MLP}\big(
    \mathrm{LN}\big(
        \tilde{\bm{\mathsf{b}}}^n
    \big)
    \big).
\label{eq:att-mlp}
\end{align}
where $\mathrm{MHA}$ denotes the multi-head attention layer \cite{vaswani2017attention},
$\mathrm{LN}$ denotes layer normalization \cite{ba2016layer}
and $\mathrm{MLP}$ denotes a feed-forward fully-connected network 
with $\mathrm{ReLU}$ activations.
We organize the skip connections following \cite{radford2019language}
to allow identity mappings \cite{he2016identity}.
This produces a set of box embeddings
$\mathcal{B}=\{\bm{\mathsf{b}}^1, \bm{\mathsf{b}}^2, \cdots, \bm{\mathsf{b}}^N\}$
(each of dimension $\lvert\bm{\mathsf{b}}\rvert=d$),
which are kept fixed during the entire episode.
We note that this is in contrast to several previous works
\cite{li2020solving,jiang2021solving}
that need to re-encode the box embeddings 
at every time step during an episode.

\subsubsection{Frontier Embedding}
\label{sec:frontier-embedding}
Besides the box embeddings $\mathcal{B}$,
in order to provide the agent with the description of the situation inside the bin, 
we additionally feed the agent with a frontier embedding 
$\bm{\mathsf{f}}$
(also of size $\vert\bm{\mathsf{f}}\vert=d$)
at each time step.
More specifically,
we define the frontier 
$\bm{\mathsf{F}}$
of a bin as the front-view of the loading structure.
For a bin of width $W$ and height $H$, 
its frontier is of size $W\times H$,
with each entry representing for that particular location 
the maximum 
$x$
coordinate of the already packed boxes.
To compensate for possible occlusions,
at each time step $t$,
we stack the last two frontiers 
$\{\bm{\mathsf{F}}_{t-1}, \bm{\mathsf{F}}_{t}\}$
($\bm{\mathsf{F}}_{t}$
denotes the frontier before the box
$(l^{s_t}, w^{s_t}, h^{s_t})$ 
has been packed into the bin)
and pass them through a convolutional network
to obtain the frontier embedding 
$\bm{\mathsf{f}_t}$.

\subsection{Decoding Sequence}
\label{sec:decoding-sequence}
We build on the pointer network 
\cite{vinyals2015pointer,bello2016neural,kool2018attention}
to model the sequence policy
$\pi^{\mathrm{s}}$
in order to allow for variable length inputs.

At time step $t$,
the task for 
the sequence policy 
$\pi^{\mathrm{s}}$
is to select the index $s_t$ of the next box to be packed
out of all unpacked boxes:
$s_t\in\{1,\ldots,N\}\setminus s_{1:t-1}$.
The embeddings for these \enquote{leftover} boxes
consistute the set
$\mathcal{B}\setminus{\bm{\mathsf{b}}^{s_{1:t-1}}}$.
Besides the leftover embeddings,
those already packed boxes 
along with the situation inside the bin
can be described by the frontier and 
subsequently the frontier embedding $\bm{\mathsf{f}}_t$.
Therefore we instantiate the encoding function
$f_{\mathcal{I}}^{\mathrm{s}}$ (Eq.\ref{eq:pi})
for the sequence policy as
\begin{align}
    \bar{\bm{\mathsf{q}}}^{\mathrm{s}}_t
=
    f^{\mathrm{s}}_{\mathcal{I}}(s_{1:t-1}, p_{1:t-1}; \bm{\theta}^{\mathrm{e}})
=
    \big\langle
        \langle
            \mathcal{B}
                \setminus{
                    \bm{\mathsf{b}}^{s_{1:t-1}} 
                }
        \rangle,
        \bm{\mathsf{f}}_{t}
    \big\rangle,
\label{eq:fs}
\end{align}
with
$\langle\rangle$
representing the operation that takes in 
a set of $d-$dimensional vectors
and return their mean vector 
which is also $d-$dimensional.
This encoding function generates 
the query vector 
$\bar{\bm{\mathsf{q}}}^{\mathrm{s}}_t$
for sequence decoding.

Following 
\cite{bello2016neural,kool2018attention},
we add a glimpse operation
via multi-head 
($M$ heads) 
attention
\cite{vaswani2017attention} 
before calculating the policy probabilities,
which has been shown to bring performance gains.
At the beginning of each episode,
from each of the box embeddings
$\bm{\mathsf{b}}^n$,
$M$ sets of fixed context vectors are obtained
via learned linear projections.
With $h=\frac{d}{M}$,
the context vectors
$\bar{\bm{\mathsf{k}}}^{n,m}$ (glimpse key of size $h$) and
$\bar{\bm{\mathsf{v}}}^{n,m}$ (glimpse value of size $h$)
for the $m^{\mathrm{th}}$ head,
as well as
$\bm{\mathsf{k}}^{n}$ (logit key of size $d$)
shared across all $M$ heads
are obtained through
\begin{align}
    \bar{\bm{\mathsf{k}}}^{n,m}
=
    \bm{\mathsf{W}}^{\bar{\bm{\mathsf{k}}},m}
    \bm{\mathsf{b}}^n,
\hspace{5pt}
    \bar{\bm{\mathsf{v}}}^{n,m}
=
    \bm{\mathsf{W}}^{\bar{\bm{\mathsf{v}}},m}
    \bm{\mathsf{b}}^n,
\hspace{5pt}
    \bm{\mathsf{k}}^{n}
=
    \bm{\mathsf{W}}^{\bm{\mathsf{k}}}
    \bm{\mathsf{b}}^n,
\label{eq:g-proj}
\end{align}
with 
$\bm{\mathsf{W}}^{\bar{\bm{\mathsf{k}}},m},
\bm{\mathsf{W}}^{\bar{\bm{\mathsf{v}}},m}
\in\mathbb{R}^{h\times d}$
and
$\bm{\mathsf{W}}^{\bm{\mathsf{k}}}\in\mathbb{R}^{d\times d}$
(in general the key and value vectors
are not required to be of the same dimensionality,
here
$\bm{\mathsf{W}}^{\bar{\bm{\mathsf{k}}},m}$ and
$\bm{\mathsf{W}}^{\bar{\bm{\mathsf{v}}},m}$ are of the same size
only for notational convenience).
The sequence query vector
$\bar{\bm{\mathsf{q}}}^{\mathrm{s}}_t$ is split into
$M$ glimpse query vectors
each of dimension
$\lvert \bar{\bm{\mathsf{q}}}^{\mathrm{s},m}_t \rvert=h$.
The compatibility vector 
$\bar{\bm{\mathsf{c}}}^m_t$ 
($\lvert \bar{\bm{\mathsf{c}}}^m_t \rvert=N$) 
is then calculated with its entries
\begin{align}
    \bar{\bm{\mathsf{c}}}^m_{t,n}
=
    \begin{cases}
        \frac{
                {\bar{\bm{\mathsf{q}}}_t^{{\mathrm{s},m}^{\top}}} 
                \bar{\bm{\mathsf{k}}}^{n,m}
             }
             {\sqrt{h}} 
& 
        \text{if} \hspace{2pt} n\centernot\in \{s_{1:t-1}\},
\\
        -\infty 
& 
        \text{otherwise}.
    \end{cases}
\end{align}
Then we can obtain the updated sequence query for the 
$m^{\mathrm{th}}$ head as the weighted sum of the glimpse value vectors
\begin{align}
    \bm{\mathsf{q}}^{\mathrm{s},m}_t
&=
    \sum_{n=1}^N
    \mathrm{softmax} (\bar{\bm{\mathsf{c}}}^m_{t})_{n}
    \cdot
    \bar{\bm{\mathsf{v}}}^{n,m}.
\end{align}
Concatenating the updated sequence query vectors 
from all $M$ heads we obtain
$\bm{\mathsf{q}}^{\mathrm{s}}_t$
($\lvert \bm{\mathsf{q}}^{\mathrm{s}}_t \rvert=d$).
Passing 
it
through another linear projection
$\bm{\mathsf{W}}^{\bm{\mathsf{q}}}
\in
\mathbb{R}^{d\times d}$,
the updated compatibility 
$\bm{\mathsf{c}}_t$ ($\lvert \bm{\mathsf{c}}_t \rvert=N$) 
is calculated by
\begin{align}
    \bm{\mathsf{c}}_{t,n}
=
    \begin{cases}
        C \cdot \mathrm{tanh} 
        \Big( 
        \frac{
                (\bm{\mathsf{W}}^{\bm{\mathsf{q}}}
                 \bm{\mathsf{q}}^{\mathrm{s}}_t)^{\top}
                \bm{\mathsf{k}}^{n}
             }
             {\sqrt{d}} 
        \Big)
& 
        \text{if} \hspace{2pt} n\centernot\in \{s_{1:t-1}\},
\\
        -\infty 
& 
        \text{otherwise},
    \end{cases}
\end{align}
where the logits are clamped between
$[-C,C]$ following
\cite{bello2016neural,kool2018attention}
as this could help exploration and 
bring performance gains
\cite{bello2016neural}. 

Finally the sequence policy can be obtained via 
\begin{align}
    \pi^{\mathrm{s}}
        \big(
            \cdot 
            \big\vert 
            f^{\mathrm{s}}_{\mathcal{I}}(s_{1:t-1}, p_{1:t-1}; \bm{\theta}^{\mathrm{e}}); 
            \bm{\theta}^{\mathrm{s}}
        \big)
=
    \mathrm{softmax}(\bm{\mathsf{c}}_t).
\label{eq:s-pi}
\end{align}
During training,
the box index $s_t$ is selected by sampling from the distribution
defined by the sequence policy
$s_t
\sim
\pi^{\mathrm{s}}
    \big(
        \cdot \big\vert f^{\mathrm{s}}_{\mathcal{I}}(s_{1:t-1}, p_{1:t-1})
    \big)
$
while during evaluation the greedy action is selected
$s_t
=
\arg\max_{a}\pi^{\mathrm{s}}
    \big(
        a \big\vert f^{\mathrm{s}}_{\mathcal{I}}(s_{1:t-1}, p_{1:t-1})
    \big)
$.

\subsection{Decoding Placement}
\label{sec:decoding-placement}

Given the box index $s_t$ selected by the sequence policy 
$\pi^{\mathrm{s}}$,
the placement policy 
$\pi^{\mathrm{p}}$
will select a placement
$p_t=(o^{s_t}, y^{s_t})$
that decides 
for this box
its orientation 
$o^{s_t}$
and its coordinate 
$y^{s_t}$
along the width dimension 
within the bin.
As discussed in Sec.\ref{sec:decomposing},
once
$y^{s_t}$
is selected,
the other two coordinates of the box
$x^{s_t}$
and 
$z^{s_t}$
can be 
directly located.

As for the encoding function 
$f_{\mathcal{I}}^{\mathrm{p}}$
for the placement policy,
it contains the processing of three kinds of embeddings:
the current box embedding $\bm{\mathsf{b}}^{s_t}$,
the leftover embeddings
$\mathcal{B}\setminus{\bm{\mathsf{b}}^{s_{1:t}}}$
and the frontier embedding $\bm{\mathsf{f}}_t$
(which is the same as for
$f_{\mathcal{I}}^{\mathrm{s}}$
since the situation inside the bin has not changed
after $s_t$ is selected).
Therefore 
$f_{\mathcal{I}}^{\mathrm{p}}$
takes the form of
\begin{align}
    \bm{\mathsf{q}}^{\mathrm{p}}_t
=
    f^{\mathrm{p}}_{\mathcal{I}}(s_{1:t}, p_{1:t-1}; \bm{\theta}^{\mathrm{e}})
=
    \big\langle
        \bm{\mathsf{b}}^{s_t}, 
        \langle
            \mathcal{B}
            \setminus{
                \bm{\mathsf{b}}^{s_{1:t}} 
            }
        \rangle,
        \bm{\mathsf{f}}_{t}
    \big\rangle.
\label{eq:fp}
\end{align}
This yields the query vector 
$\bm{\mathsf{q}}^{\mathrm{p}}_t$
for placement decoding.

Unlike the sequence policy 
$\pi^{\mathrm{s}}$
which is selecting over a fixed set of candidate boxes
for each time step during an episode,
the candidate pool for the placement policy 
$\pi^{\mathrm{p}}$
is changing from time step to time step, 
which makes it not quite suitable to model
$\pi^{\mathrm{p}}$
using another pointer network
\cite{vinyals2015pointer}
with nonparametric 
$\mathrm{softmax}$.

Therefore,
we deploy a placement network
(parameterized by $\bm{\theta}^{\mathrm{p}}$)
that takes in the query vector
$\bm{\mathsf{q}}^{\mathrm{p}}_t$
and output action probabilities
with standard parametric
$\mathrm{softmax}$.
More specifically,
the placement network 
$\bm{\theta}^{\mathrm{p}}$
outputs logits 
$\bar{\bm{\mathsf{l}}}^{\mathrm{p}}_t
=
\bm{\theta}^{\mathrm{p}}
(\bm{\mathsf{q}}^{\mathrm{p}}_t)$ 
of size $O\times W$.
If a placement 
$(o_i,y_j)$ is infeasible
that after being placed
the current box 
will cut through walls of the container,
its corresponding entry in the logit vector 
$\bar{\bm{\mathsf{l}}}^{\mathrm{p}}_{t,i\times j}$
will be masked as
$-\infty$,
while all other feasible entries will be processed as 
$C\cdot\mathrm{tanh}
(\bar{\bm{\mathsf{l}}}^{\mathrm{p}}_{t,\cdot})$.
This processed logit vector
$\bm{\mathsf{l}}^{\mathrm{p}}_t$
is then used to give out action probabilities for placement selection
\begin{align}
    \pi^{\mathrm{p}}
        \big(
            \cdot 
            \big\vert 
            f^{\mathrm{p}}_{\mathcal{I}}(s_{1:t}, p_{1:t-1}; \bm{\theta}^{\mathrm{e}});
            \bm{\theta}^{\mathrm{p}}
        \big)
=
    \mathrm{softmax}(
        \bm{\mathsf{l}}^{\mathrm{p}}_t
    ).
\end{align}
As in sequence selection,
the placement 
$p_t$
is sampled
during training
while
chosen greedily
during evaluation.

\subsection{Policy Gradient}
So far we have been discussing that given an input set
$\mathcal{I}$,
how the sequence agent 
$\pi^{\mathrm{s}}$
and the placement agent
$\pi^{\mathrm{p}}$
cooperatively yield a solution configuration
$\mathcal{C}_{\pi}
=
\{s_1, p_1, s_2, p_2, \ldots, s_N, p_N\}$
in an interleaving manner over $N$ time steps
($\pi$ denotes the aggregation of
$\pi^{\mathrm{s}}$
and
$\pi^{\mathrm{p}}$).
In order to train this system such that
$\pi^{\mathrm{s}}$
and
$\pi^{\mathrm{p}}$
could cooperatively maximize the final utility
$r_{\mathrm{u}}\in[0,1]$,
which is equivalent to minimizing the cost 
$c(\mathcal{C}_{\pi})
=
1-r_{\mathrm{u}}(\mathcal{C}_{\pi})$,
we define the overall loss function as
the expected cost
of
the configurations 
generated by following
$\pi$:
\begin{align}
    \mathcal{J}(\bm{\theta} \vert \mathcal{I})
=
    \mathbb{E}_{\mathcal{C}_{\pi}\sim\pi_{\bm{\theta}}}
    \left[
        c(\mathcal{C}_{\pi} \vert \mathcal{I})
    \right],
\end{align}
which is optimized by following the 
$\mathrm{REINFORCE}$ gradient estimator
\cite{williams1992simple}
\begin{align}
    &\nabla_{\bm{\theta}}
    \mathcal{J}(\bm{\theta} \vert \mathcal{I})
=
    \mathbb{E}_{\mathcal{C}_{\pi}\sim\pi_{\bm{\theta}}}
    \Big[
        \Big(
            c(\mathcal{C}_{\pi} \vert \mathcal{I})
            - 
            b(\mathcal{I})
        \Big)
    \cdot
\notag\\&
    \sum_{t=1}^{N}
        \Big(
        \nabla_{\bm{\theta}^{\mathrm{e},\mathrm{s}}}
        \log \pi^{\mathrm{s}} 
        (
            s_t; 
            \bm{\theta}^{\mathrm{e},\mathrm{s}}
        )
        +    
        \nabla_{\bm{\theta}^{\mathrm{e},\mathrm{p}}}
        \log \pi^{\mathrm{p}}
        (
            p_t; 
            \bm{\theta}^{\mathrm{e},\mathrm{p}}
        )
        \Big)
    \Big].
\label{eq:pg}
\end{align}
We note that to avoid cluttering,
we consume the dependency of 
$\pi^{\mathrm{s}}$
on
$f^{\mathrm{s}}_{\mathcal{I}}(s_{1:t-1}, p_{1:t-1};\bm{\theta}^{\mathrm{e}})$
into
$\bm{\theta}^{\mathrm{e}}$
and use
$\bm{\theta}^{\mathrm{e},\mathrm{s}}$
to denote the aggregation of
$\bm{\theta}^{\mathrm{e}}$
and
$\bm{\theta}^{\mathrm{s}}$;
the same as for 
$\pi^{\mathrm{p}}$.

For the baseline function
$b(\mathcal{I})$
in the above equation,
we use the greedy rollout baseline proposed by 
\cite{kool2018attention}
as this has been shown to give superior performance.
More specifically,
the parameters of the best model $\bm{\theta}^{\mathrm{BL}}$ 
encountered during training is used to 
roll out greedily on the training input $\mathcal{I}$ 
to give out 
$b(\mathcal{I})$;
at the end of every epoch $i$,
the model $\bm{\theta}^{\mathrm{BL}}$
will be compared against the current model $\bm{\theta}_i$
and will be replaced if the improvement of
$\bm{\theta}_i$
over
$\bm{\theta}^{\mathrm{BL}}$
is significant according to a paired t-test ($\alpha=5\%$).
More details can be found in
\cite{kool2018attention}.

\subsection{Prioritized Oversampling}

Based on the observations from preliminary experiments,
we suspect that the placement policy
could have a more local effect 
than the sequence policy
on the overall quality 
of the resulting configuration:
we found that in hard testing scenarios,
a sub-optimal early sequence selection made by 
$\pi^{\mathrm{s}}$ 
would make it challenging for the overall system 
to yield a satisfactory solution configuration; 
while given a box to place, 
the effect of a sub-optimal decision of 
$\pi^{\mathrm{p}}$
usually do not propagate far. 
Another obsevation that leads to this suspision
is that those testing samples 
on which the trained system perform poorly  
are usually those that have a more strict
requirement on the sequential order of the input boxes. 

However,
in our problem setup,
the agent would only receive an evaluation signal 
upon the completion of a full episode,
which involves 
subsequently rolling out 
$\pi^{\mathrm{s}}$
and
$\pi^{\mathrm{p}}$
over multiple time steps,
with sequence selection and 
placement selection interleavingly 
conditioned and dependent on each other.
This makes it not quite suitable to use schemes
such as utilizing a counterfactual baseline
\cite{foerster2018counterfactual}
to address multi-agent credit assignment,
especially that in our case
marginalizing out the effect of a sequence action $s_t$
is not straightforward and could be expensive.



So instead,
inspired by prioritized experience replay
\cite{schaul2015prioritized}
for off-policy learning,
we propose prioritized oversampling (PO) 
as a general technique
to speed up on-policy training.
The general idea of
prioritized oversampling is 
to give the agent a second chance of learning
those hard examples on which 
it had performed poorly during regular training.
Following the above discussions,
we focus on finding better sequence selections
for those hard samples
during the second round of learning.
To achieve this we first follow 
a beam-search-based scheme;
but we later found that a much simpler
re-learning procedure could bring performance gains
on par with the rather 
complicated beam-search-based training,
therefore we choose the simple alternative 
as our final design choice.

Prioritized oversampling   
is carried out in two stages:
collect and re-learn.
First,
during training,
when learning on a batch 
$\mathcal{B}$
of batch size
$B$,
the advantages of this batch are calculated as
$
c(\mathcal{C}_{\pi} \vert \mathcal{I})
- 
b(\mathcal{I})
$
(Eq.\ref{eq:pg}).
We note that the lower the advantage 
the better is the configuration
$\mathcal{C}_{\pi}$,
since here we use the notion of 
costs instead of rewards. 
From those training samples, 
a PO batch 
$\mathcal{B}^{\mathrm{PO}}$
of size
$B^{\mathrm{PO}}$
(usually 
$B^{\mathrm{PO}} \ll B$)
is then collected
that contains those with the highest advantage values
hence the worst solution configurations.
After a total number of 
$B$
PO samples have been collected
(e.g. after
$\nicefrac{B}{B^{\mathrm{PO}}}$
steps of normal training), 
the second step of PO re-learning 
can be carried out.
We note that it might not 
be necessary to conduct normal training
and PO re-learning using the same batch size, 
while we found out empirically
that this lead to more stable training dynamics
especially when learning with optimizers
with adaptive learning rates
\cite{kingma2015adam}.
We also deploy a separate optimizer for PO re-learning and 
found this to be crucial to stabilize learning,
as the statistics of the PO batches 
could differ dramatically 
from the normal training batches.

As for the PO re-learning,
we experiment with two types of 
learning strategies:
beam-search-based training 
and normal training. 
In beam-search-based training
with a beam size of 
$K$,
for each PO sample,
$K$ instead of $1$
solution configurations will be generated 
by the end of an episode.
More specifically,
at 
$t=1$,
$\pi^{\mathrm{s}}$
will sample 
$K^2$
box indices
$\{s_{1,1}, s_{1,2}, \cdots, s_{1,K^2}\}$.
These will all be given to
$\pi^{\mathrm{p}}$
which will select a placement for each 
$s_{1,k}$,
out of which
$K$
candidates with the highest
$\pi^{\mathrm{s}}(s_{1,k})
\pi^{\mathrm{p}}(p_{1,k})$ 
will be selected to be passed 
as the starting configuration
to the next time step.
In all time steps
$t>1$,
$\pi^{\mathrm{s}}$
will receive
$K$
candidate starting configurations,
and it will sample for each of these
$K$
configurations
$K$
next box indices.
These 
$K^2$
box indices will be passed to
$\pi^{\mathrm{p}}$
to select one placement for each.
Pruning will again be conducted according to
$\prod_{\tau=1}^{t}
\pi^{\mathrm{s}}(s_{\tau})
\pi^{\mathrm{p}}(p_{\tau})$
from which 
$K$
candidates are left.
From preliminary experiments with this setup,
we found that those 
$K$
final trajectories picked out by beam-search 
do not necessarily yield better performance,
we suspect that the reason is that 
the pruning criteria used here does not necessarily
translate to good utility.
We then adjust the algorithm that after the
$K$
configurations have been generated 
at the end of episodes,
we only choose the one 
with the best advantage to train the model.
We found that this give better performance.
However,
we found that a much simpler training scheme,
that instead of conducting
training-time beam-search,
simply conduct
regular training
on the PO samples 
during the re-learning phase could already bring
performance gains,
albeit being much simpler 
in concept and in implementation.
Therefore we choose the latter strategy 
as the final design choice.

We note that in order to compensate for oversampling,
during the normal training of batch
$\mathcal{B}$, 
we scale down the losses of 
each of those samples
that are collected into
$\mathcal{B}^{PO}$ by
$\eta$;
while during PO re-learning,
the coefficient 
for those PO samples
is set to
$1-\eta$.
$\eta$ is calculated by tracking 
a set of running statistics. 
We denote the samples in 
$\mathcal{B}$
with positive advantages hence 
relatively inferior performance as
$\mathcal{B}^{+}$,
and denote its size as
$B^{+}$.
Then we keep track of the running statistics
(with momentum $\beta$)
of the sample size
$B^{+}$
and of the average advantages of
$\mathcal{B}^{+}$,
denoted as
$B^{+}_{\beta}$
and
$a^{+}_{\beta}$
respectively,
which are updated by: 
$a^{+}_{\beta}
=
    \nicefrac{(
    \beta \cdot B^{+}_{\beta} \cdot a^{+}_{\beta}
    +
        (1-\beta) \cdot B^{+} \cdot a^{+}
    )}
    {(
        \beta \cdot B^{+}_{\beta} 
    +
        (1-\beta) \cdot B^{+}
    )}
$,
$B^{+}_{\beta}
=
    \beta \cdot B^{+}_{\beta} 
+
    (1-\beta) \cdot B^{+}
$.
We conduct the same running statistics tracking for the PO batches
$\mathcal{B}^{\mathrm{PO}}$
from which we get
$a^{\mathrm{PO}}_{\beta}$.
Then we get the coefficient 
$\eta
=
\max(
\min(1 - \nicefrac{a^{+}_{\beta}} {a^{\mathrm{PO}}_{\beta}}, 
     1.0),
0.5)$.


\section{Experiments}
\label{sec:exp}

\subsection{Experimental Setup}

When conducing experiments to validate our proposed method,
we do not find it straightforward 
to compare against previous works, 
as the datasets used in previous evaluations 
vary from one to another.
The main varying factors are 
the size of the bin, 
the number of boxes to pack,
the range of the size of the boxes,
as well as how the box dimensions are sampled.
In order to offer as a valid benchmark,
in this paper we conduct 
ablations as well as
careful studies 
against a variety of previous state-of-the-art methods on BPP,
comparing to each according to its specific setups.
Unless otherwise mentioned,
we run
$2$
trainings 
under
$2$
non-tuned random seeds
for each training configuration.


Across all experiments,
we use the Adam optimizer
\cite{kingma2015adam}
with a learning rate of
$1\mathrm{e}{-4}$
and a batch size of 
$B=1024$,
and each epoch contains
$1024$
such minibatches.
After every epoch,
evaluation is conducted 
on a fixed set of 
validation data of size
$1\mathrm{e}{4}$.
The plots we present in this section
are all of the performance of these evaluations during training,
while the final testing results presented in tables 
are on testing data 
of size
$1\mathrm{e}{4}$
that are drawn independently 
from the training/validation data,
and the testing results 
of the best agent out of the $2$ runs 
are presented in the tables.
A fixed training budget of
$128$
epochs is set for our full model.
For those experiments with 
prioritized oversampling (PO),
a PO batch of size
$B^{\mathrm{PO}}=16$
is collected from each training batch,
which means that a full batch ready for PO re-learning
would be gathered every 
$\nicefrac{B}{B^{\mathrm{PO}}}=64$
training steps.
This means that after an epoch of
$1024$
training steps,
$\nicefrac{1024}{64}=16$
more PO training steps would have been conducted
than setups without PO.
Over a total number of 
$128$
epochs this sums up to
$128\cdot16=2\cdot1024$,
equals to
$2$
full epochs of more training steps.
Thus to ensure a fair comparison,
we set the total number of epochs 
for those setups without PO to 
$130$.

The raw box dimensions are normalized 
by the maximum size of the box edges
before fed into the deep network.
As for the network,
the initial linear layer
$\mathrm{FF}$
contains 
$128$
hidden units,
which is followed by
$3$
multi-head 
(with $M=8$ heads)
self-attention layers 
with an embedding size of
$d=128$.
The costs 
$c$
are scaled up by
$3.0$
before used to calculate losses.
Other important hyperparameters are
$C=10$,
$\beta=0.95$.

\begin{figure}[t]
	\centering
	\begin{subfigure}{0.87\columnwidth}
		\includegraphics[width=\columnwidth]{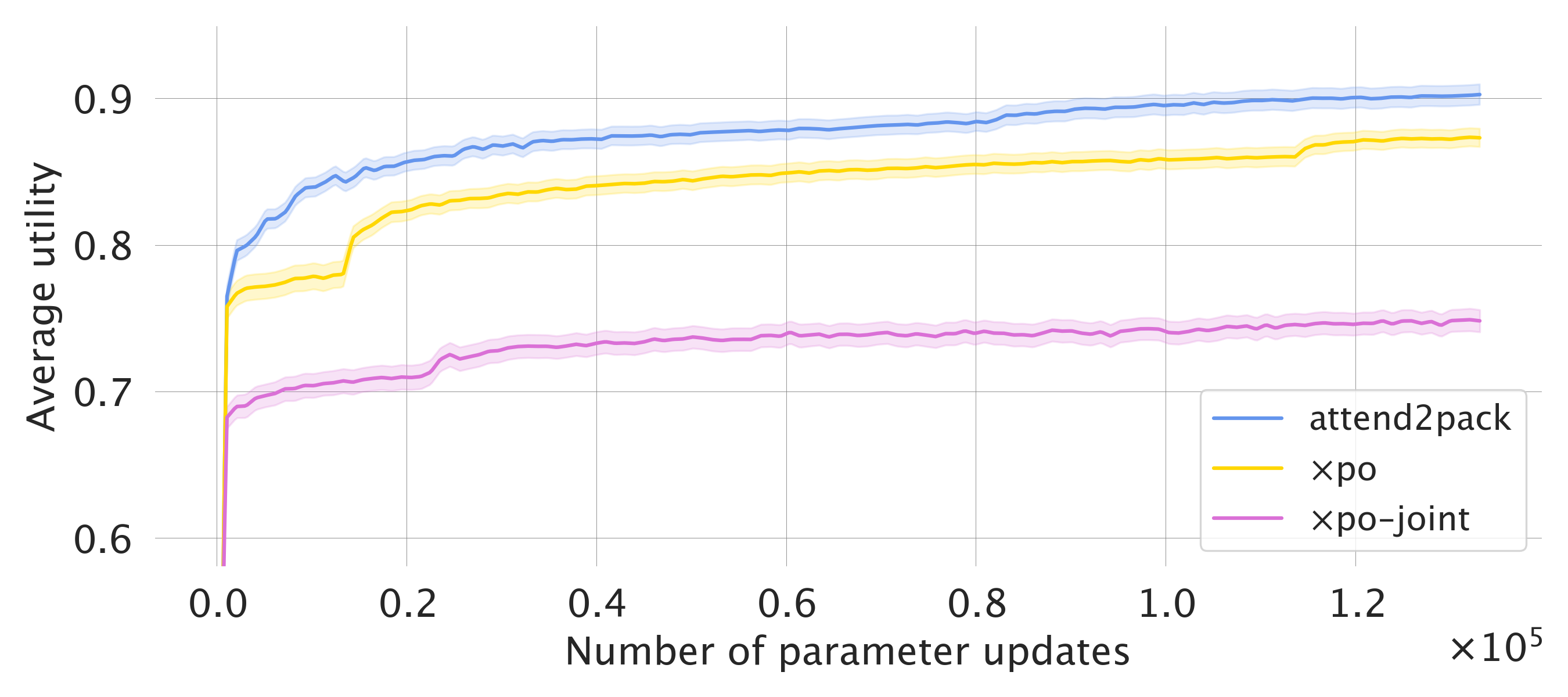}
\vspace{-15pt}
		\caption{
Average utility obtained in evaluation during training,
each plot shows mean $\pm\hspace{3pt}0.1\hspace{2pt}\cdot$ standard deviation.
		}
		\label{fig:cut-rr-ablation-visual-a}
	\end{subfigure}
	\begin{subfigure}{0.25\columnwidth}
		\includegraphics[width=\columnwidth]{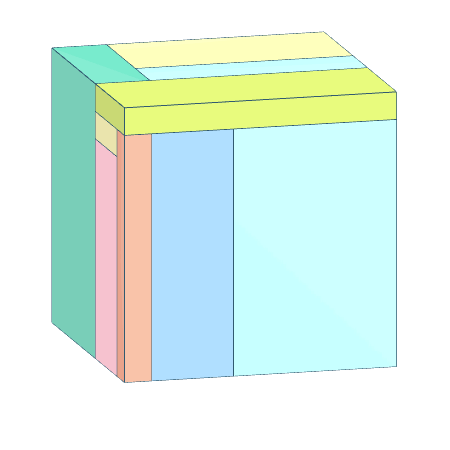}
\vspace{-20pt}
		\caption{
			$\mathrm{attend2pack}$
		}
		\label{fig:cut-rr-ablation-visual-b}
	\end{subfigure}
	\begin{subfigure}{0.25\columnwidth}
		\includegraphics[width=\columnwidth]{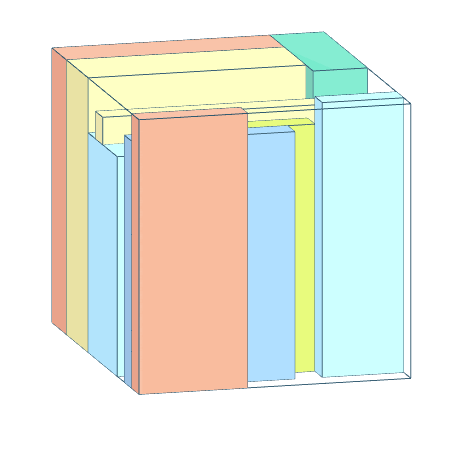}
\vspace{-20pt}
		\caption{
			$\text{\texttimes}\mathrm{po}$
		}
		\label{fig:cut-rr-ablation-visual-c}
	\end{subfigure}
	\begin{subfigure}{0.25\columnwidth}
		\includegraphics[width=\columnwidth]{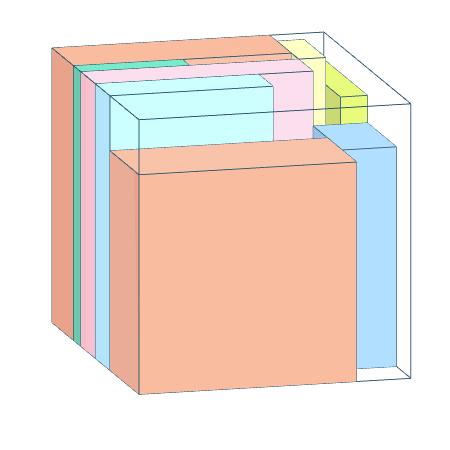}
\vspace{-20pt}
		\caption{
			$\text{\texttimes}\mathrm{po}
				\mbox{-}
				\mathrm{joint}$
		}
		\label{fig:cut-rr-ablation-visual-d}
	\end{subfigure}
\vspace{-5pt}
	\caption{
Plots 
(Fig.\ref{fig:cut-rr-ablation-visual-a}) 
and visualizations 
(Fig.\ref{fig:cut-rr-ablation-visual-b}-\ref{fig:cut-rr-ablation-visual-d})
for Sec.\ref{sec:cut-rr-ablation},
on dataset generated by 
cutting $10\times10\times10$ bin into $10$ boxes.
	}
	\label{fig:cut-rr-ablation-visual}
\end{figure}

\subsection{Ablation Study on Action Space}
\label{sec:cut-rr-ablation}
We first conduct an ablation study to evaluate 
the effectiveness of the action space decomposition,
as well as the effect of utilizing PO.
For this experiment,
we use the data setup as in 
\cite{laterre2018ranked},
where boxes are generated by cutting a 
bin of size
$10\times10\times10$,
and the resulting box edges
could take any integer in the range
$[1\twodots10]$.
\cite{laterre2018ranked}
conducted experiments
by cutting the aforementioned bin into
$10$,
$20$,
$30$,
$50$
boxes.
Here we conduct ablations on the 
$10$
boxes dataset,
and a full comparison 
to this state-of-the-art algorithm 
\cite{laterre2018ranked}
will be presented in
Sec.\ref{sec:cut-rr}.

Our full method is denoted
$\mathrm{attend2pack}$,
which we compare against a
$\text{\texttimes}\mathrm{po}$
agent trained without utilizing PO,
and a
$\text{\texttimes}\mathrm{po}
\mbox{-}
\mathrm{joint}$
agent directly trained on the joint action space,
with one 
$\mathrm{softmax}$ 
policy network directly outputting
action probabilities over the full action space.
The plots of evaluation during training are shown in 
Fig.\ref{fig:cut-rr-ablation-visual-a},
which shows that
decomposing the 
joint combinatorial action space 
does boost learning,
and PO helps to further improve training efficiency.
Packing results
are also visualized in 
Fig.\ref{fig:cut-rr-ablation-visual-b}-\ref{fig:cut-rr-ablation-visual-d}.

\begin{figure}[t]
	\centering
	\begin{subfigure}{0.87\columnwidth}
		\includegraphics[width=\columnwidth]{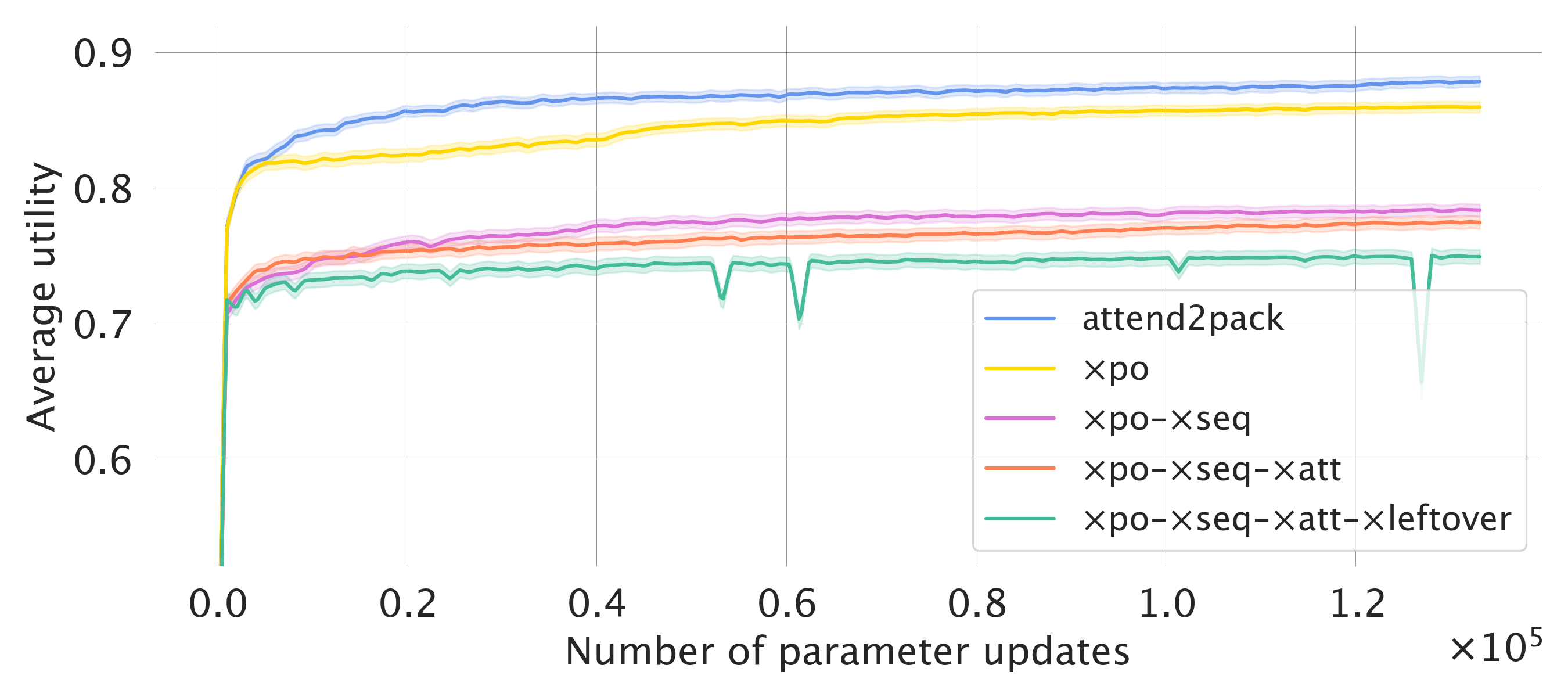}
\vspace{-15pt}
		\caption{
Average utility obtained in evaluation during training,
each plot shows mean $\pm\hspace{3pt}0.1\hspace{2pt}\cdot$ standard deviation.
		}
		\label{fig:random-online-ablation-visual-a}
	\end{subfigure}

	\begin{subfigure}{0.32\columnwidth}
		\includegraphics[width=\columnwidth]{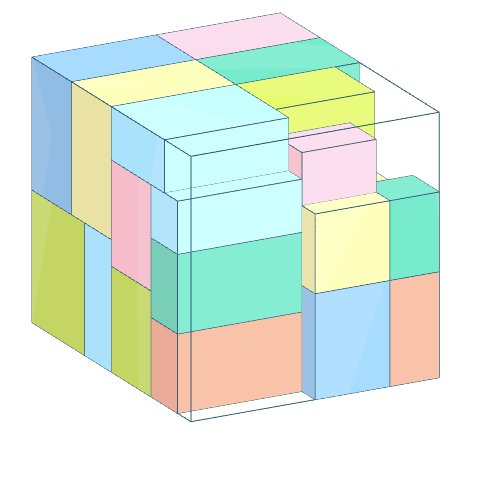}
\vspace{-15pt}
		\caption{
			$\mathrm{attend2pack}$
		}
		\label{fig:random-online-ablation-visual-b}
	\end{subfigure}
	\begin{subfigure}{0.33\columnwidth}
		\includegraphics[width=\columnwidth]{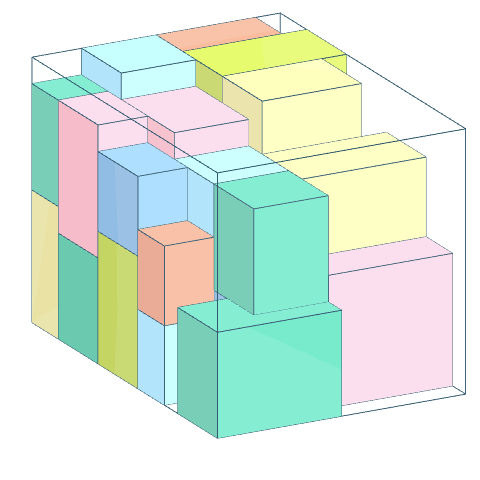}
\vspace{-15pt}
		\caption{
			$\text{\texttimes}\mathrm{po}
				\mbox{-}
				\text{\texttimes}\mathrm{seq}
				\mbox{-}
				\text{\texttimes}\mathrm{att}$
			$\mbox{-}
				\text{\texttimes}\mathrm{leftover}$
		}
		\label{fig:random-online-before}
	\end{subfigure}
	\begin{subfigure}{0.33\columnwidth}
		\includegraphics[width=\columnwidth]{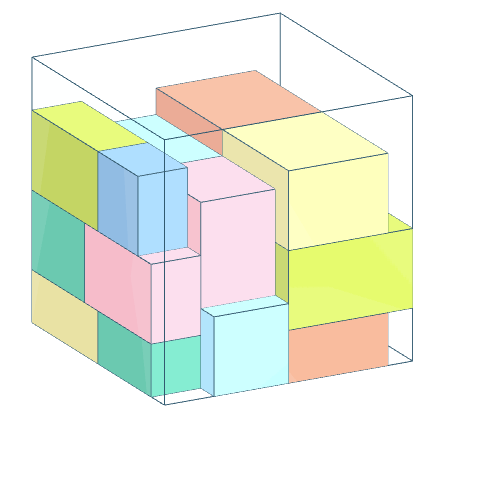}
\vspace{-15pt}
		\caption{
			$\text{\texttimes}\mathrm{po}
				\mbox{-}
				\text{\texttimes}\mathrm{seq}
				\mbox{-}
				\text{\texttimes}\mathrm{att}$
			$\mbox{-}
				\text{\texttimes}\mathrm{leftover}$
			cut
		}
		\label{fig:random-online-after}
	\end{subfigure}
	\caption{
Results for Sec.\ref{sec:random-online-ablation},
on dataset generated by randomly sampling
the edges of
$24$
boxes
from
$[2\twodots5]$,
with a bin of 
$W\times H=10\times10$.
Fig.\ref{fig:random-online-after}
shows the packing configuration of 
Fig.\ref{fig:random-online-before}
after being processed to compare 
with palletizing in onlin-BPP,
where the items at least partially outside of the 
$10\times10\times10$
bin is removed,
after which the packing is rotated
such that the rear wall of the bin is now 
at the bottom of the pallet. 
	}
	\label{fig:random-online-ablation-visual}
\end{figure}

\begin{figure}[t]
	\centering

	\begin{subfigure}{0.23\columnwidth}
		\includegraphics[width=\columnwidth]{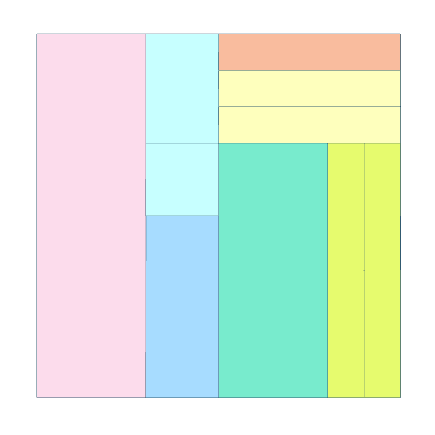}
		\caption{$10$ items}
	\end{subfigure}
	\begin{subfigure}{0.23\columnwidth}
		\includegraphics[width=\columnwidth]{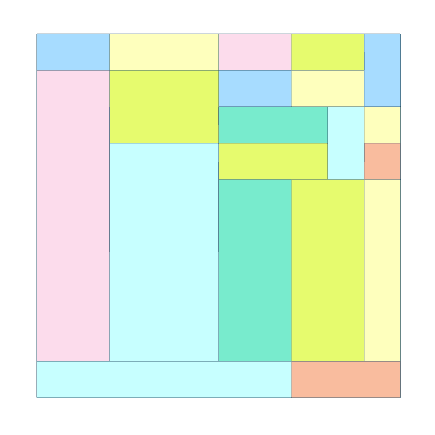}
		\caption{$20$ items}
	\end{subfigure}
	\begin{subfigure}{0.23\columnwidth}
		\includegraphics[width=\columnwidth]{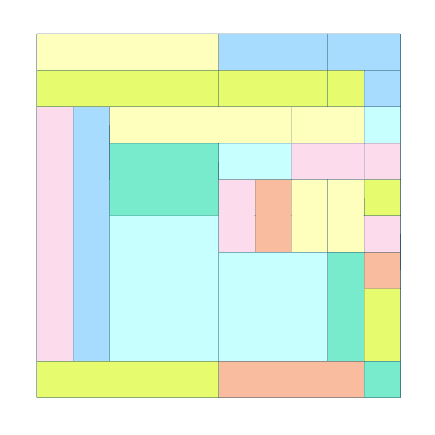}
		\caption{$30$ items}
	\end{subfigure}
	\begin{subfigure}{0.23\columnwidth}
		\includegraphics[width=\columnwidth]{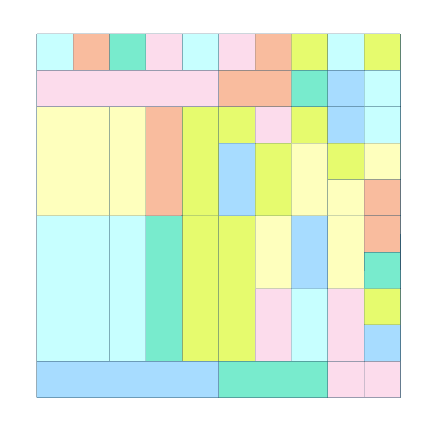}
		\caption{$50$ items}
	\end{subfigure}


	\begin{subfigure}{0.24\columnwidth}
		\includegraphics[width=\columnwidth]{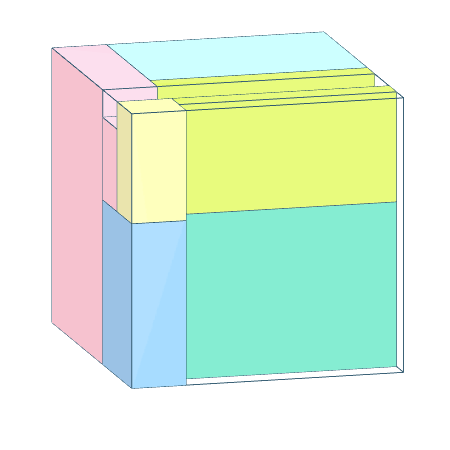}
		\caption{$10$ items}
	\end{subfigure}
	\begin{subfigure}{0.24\columnwidth}
		\includegraphics[width=\columnwidth]{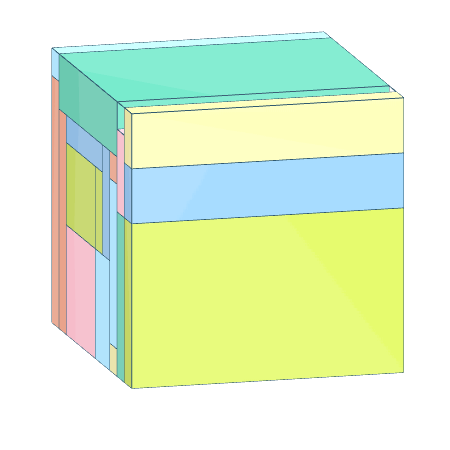}
		\caption{$20$ items}
	\end{subfigure}
	\begin{subfigure}{0.24\columnwidth}
		\includegraphics[width=\columnwidth]{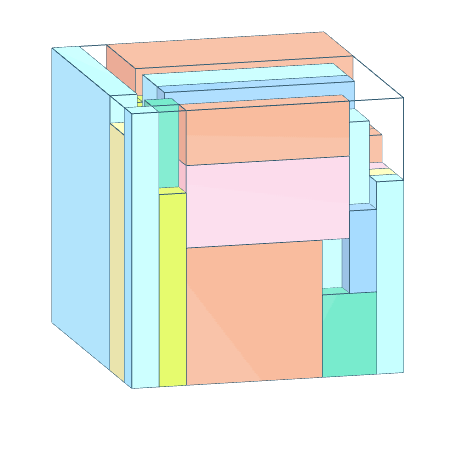}
		\caption{$30$ items}
	\end{subfigure}
	\begin{subfigure}{0.24\columnwidth}
		\includegraphics[width=\columnwidth]{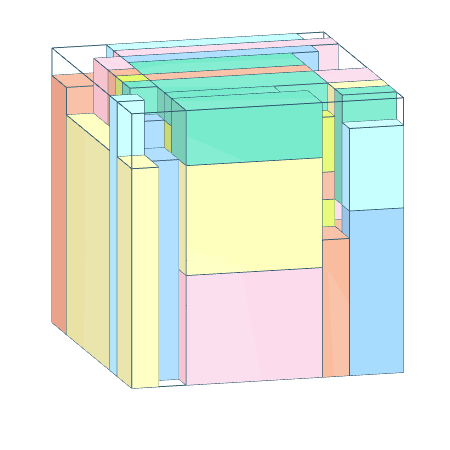}
		\caption{$50$ items}
	\end{subfigure}

	\caption{
Visualizations for Sec.\ref{sec:cut-rr},
on dataset generated by cutting 
a square of $10\times10$ ($2\mathrm{D}$) 
or a bin of $10\times10\times10$ ($3\mathrm{D}$)
into
$10$,
$20$,
$30$,
$50$
items,
with the box edges ranging within
$[1\twodots10]$.
	}
	\label{fig:cut-rr}
\end{figure}
\begin{table}[t]
  \caption{
    Performance comparison on the cut dataset
    with ranked reward
    ($\mathrm{RR}$)
    \cite{laterre2018ranked}.
    The first two columns show results in
    $r_{\mathrm{RR}}$,
    while the last column gives
    $r_{\textrm{u}}$
    of our method to ease comparisons
    of future works \footnotemark.
  }
  \label{tab:cut-rr}
  \begin{center}
    \begin{scriptsize}
      \begin{sc}
        \begin{tabular}{c|c|c|c}
          \toprule
          \multicolumn{1}{l}{} & RR ($r_{\mathrm{RR}}$) & Attend2Pack ($r_{\mathrm{RR}}$)    & Attend2Pack ($r_{\mathrm{u}}$) \\
          \midrule
          2D-10                & $0.953\pm0.027$        & $\bm{0.995}\pm0.015$ & $\bm{0.991}\pm0.028$   \\
          2D-20                & $0.948\pm0.020$        & $\bm{0.997}\pm0.012$ & $\bm{0.994}\pm0.022$   \\
          2D-30                & $0.954\pm0.015$        & $\bm{0.999}\pm0.006$ & $\bm{0.999}\pm0.011$   \\
          2D-50                & $0.960\pm0.016$        & $\bm{1.000}\pm0.000$ & $\bm{1.000}\pm0.000$    \\
          \midrule
          3D-10                & $0.903\pm0.060$        & $\bm{0.953}\pm0.042$ & $\bm{0.932}\pm0.060$  \\
          3D-20                & $0.850\pm0.032$        & $\bm{0.911}\pm0.034$ & $\bm{0.872}\pm0.046$  \\
          3D-30                & $0.844\pm0.030$        & $\bm{0.904}\pm0.033$ & $\bm{0.863}\pm0.044$  \\
          3D-50                & $0.838\pm0.034$        & $\bm{0.926}\pm0.024$ & $\bm{0.893}\pm0.032$  \\
          \bottomrule
        \end{tabular}
      \end{sc}
    \end{scriptsize}
  \end{center}
\end{table}

\footnotetext{We note here the performance of
  $\mathrm{RR}$
  are without MCTS simulations,
  as these results are presented numerically in
  \cite{laterre2018ranked},
  while the results with MCTS simulations
  are only presented in box plots.
  In the box plots,
  the RR method with MCTS simulation is compared against
  plain MCTS
  \cite{browne2012survey},
  the Lego heuristic
  \cite{hu2017solving},
  and the GUROBI solver
  \cite{gurobi2018gurobi},
  where RR shows superior performance against all of these methods.
  By roughly inferring the numerical results from the box plots
  we can conclude that our method still outperforms
  RR with MCTS simulations except for
  the scenario in $3\mathrm{D}$ with $10$ boxes.}

\begin{figure}[t]
	\centering
	\begin{subfigure}{0.87\columnwidth}
		\includegraphics[width=\columnwidth]{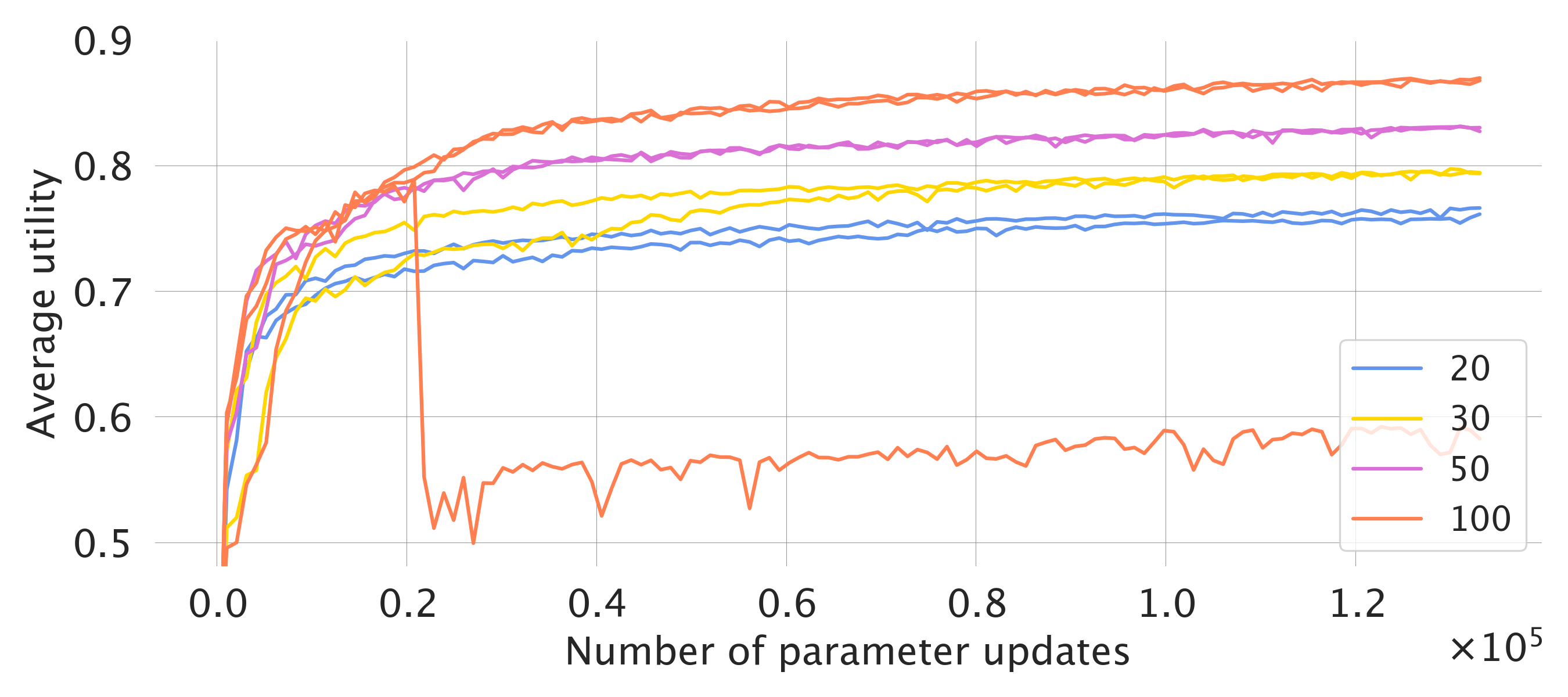}
		\caption{
Average utility obtained in evaluation during training,
here plots each individual run for each configuration.
			}
		\label{fig:random-cql-visual-a}
	\end{subfigure}
	\begin{subfigure}{0.3\columnwidth}
		\includegraphics[width=\columnwidth]{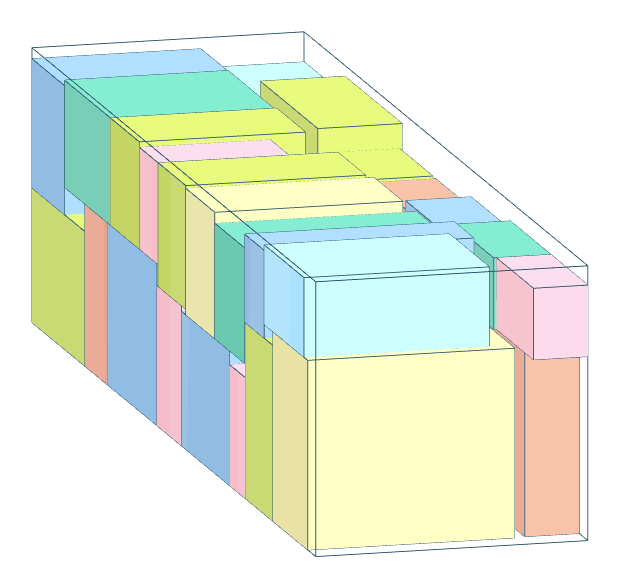}
		\caption{$30$ items}
		\label{fig:random-cql-visual-b}
	\end{subfigure}
	\begin{subfigure}{0.3\columnwidth}
		\includegraphics[width=\columnwidth]{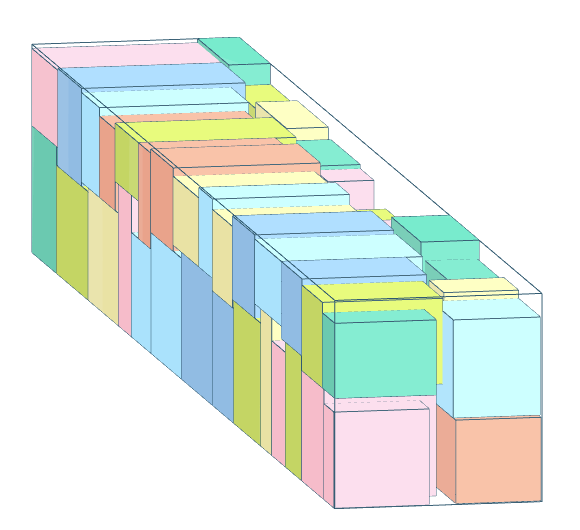}
		\caption{$50$ items}
		\label{fig:random-cql-visual-c}
	\end{subfigure}
	\begin{subfigure}{0.3\columnwidth}
		\includegraphics[width=\columnwidth]{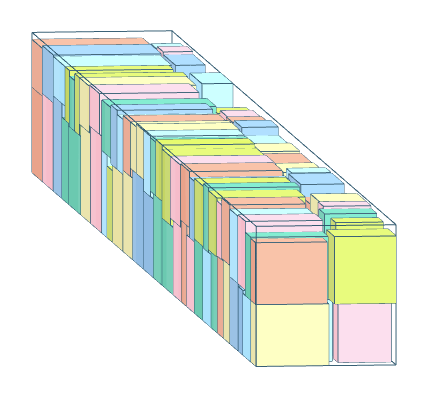}
		\caption{$100$ items}
		\label{fig:random-cql-visual-d}
	\end{subfigure}
	\caption{
Plots (Fig.\ref{fig:random-cql-visual-a})
and visualizations (Fig.\ref{fig:random-cql-visual-b}-\ref{fig:random-cql-visual-d})
for Sec.\ref{sec:random-cql},
on dataset generated by randomly sampling
the edges of
$20$,
$30$,
$50$,
$100$
boxes
from
$[20\twodots80]$,
with a bin of 
$W\times H=100\times100$.
	}
	\label{fig:random-cql-visual}
\end{figure}
\begin{table}[t]
  \caption{
  Comparisons in 
  Sec.\ref{sec:random-cql}
  on $r_{\mathrm{u}}$
  against
  GA\cite{wu2010three},
  EP\cite{crainic2008extreme},
  MTSL\cite{duan2019multi},
  CQL\cite{li2020solving} and
  MM\cite{jiang2021solving}.
  }
  \label{tab:random-cql}
  \begin{center}
    \begin{scriptsize}
      \begin{sc}
        \begin{tabular}{ccccccc}

          \toprule
                & GA      & EP      & MTSL    & CQL     & MM      & Attend2Pack                       \\
          \midrule
          $20$  & $0.683$ & $0.627$ & $0.624$ & $0.670$ & $0.718$ & $\bm{0.767}\pm0.036$ \\
          $30$  & $0.662$ & $0.638$ & $0.601$ & $0.693$ & $0.755$ & $\bm{0.797}\pm0.032$ \\
          $50$  & $0.659$ & $0.663$ & $0.553$ & $0.736$ & $0.813$ & $\bm{0.831}\pm0.023$ \\
          $100$ & $0.624$ & $0.675$ & $-$     & $-$     & $0.844$ & $\bm{0.870}\pm0.016$ \\
          \bottomrule
        \end{tabular}
      \end{sc}
    \end{scriptsize}
  \end{center}
\end{table}


\subsection{Ablation Study in the Context of Online BPP}
\label{sec:random-online-ablation}
We conduct another set of ablation studies
in the context of online-BPP
against the state-of-the-art
learning-based solution in this domain
\cite{zhao2020online}.
As discussed in the introduction,
in online-BPP,
the agent is only aware of 
the dimensions of the box at hand.
Since our algorithm targets the offline-BPP problem,
we strip our method down to the strict online setting
and also along the way
evaluate the effects of several components of our proposed method.

\cite{zhao2020online}
conducted experiments 
with boxes generated 
by cut
and by random sampling.
As comparing to their cut experiments is 
not easily feasible,
we conduct experiments using 
their random data-generation scheme:
each box edge is sampled from the integers in the range of
$[2\twodots5]$,
while the bin is of size
$10\times10\times10$.
Their evaluation criteria is 
to count the average number of boxes 
that can be placed inside the bin,
which is a bit different from our training scheme.
Thus we compare the final results against theirs
in the following way:
based on the total volume of the bin 
and the expected volume of each box, 
we can calculate that 
the maximum expected number of boxes that can be packed is
$24$. 
We thus set
$N=24$
when generating datasets.
After training, 
we remove all the boxes that are 
at least partially
out of the 
$10\times10\times10$
bin and count the number of boxes 
still left inside of the bin 
to compare to 
\cite{zhao2020online}.

The training configurations under this study are:
\begin{itemize}
    \item
$\mathrm{attend2pack}$:
Our full method;
    \item
$\text{\texttimes}\mathrm{po}$:
Our full method but without PO;
    \item
$\text{\texttimes}\mathrm{po}
\mbox{-}
\text{\texttimes}\mathrm{seq}$:
No PO training and no sequence policy learning.
While this is very similar to the online-BPP, 
since self-attention are used to calculate the box embeddings
$\mathcal{B}$,
each individual box embedding
$\bm{\mathrm{b}}^n$
still carries information about other boxes;
    \item
$\text{\texttimes}\mathrm{po}
\mbox{-}
\text{\texttimes}\mathrm{seq}
\mbox{-}
\text{\texttimes}\mathrm{att}$:
On top of the former setup,
we replace the attention module in 
$\bm{\theta}^{\mathrm{e}}$
with a residual network 
with approximately the same number of parameters.
But still this setup cannot be regarded 
as a strict online setting,
since the encoding function for placement
$f_{\mathcal{I}}^{\mathrm{p}}$
(Eq.\ref{eq:fp})
contains the leftover embeddings
which still gives information 
about the upcoming boxes.
    \item
$\text{\texttimes}\mathrm{po}
\mbox{-}
\text{\texttimes}\mathrm{seq}
\mbox{-}
\text{\texttimes}\mathrm{att}
\mbox{-}
\text{\texttimes}\mathrm{leftover}$:
A strict online-BPP setup where
the leftover embeddings are removed from 
$f_{\mathcal{I}}^{\mathrm{p}}$
from the former setup.
\end{itemize}

The results of this ablation are ploted in 
Fig.\ref{fig:random-online-ablation-visual-a}.
We can see that both 
the leftover embeddings
and the attention module brings performance gains.
It can also be concluded from the
big performance gap between
$\text{\texttimes}\mathrm{po}$
and
$\text{\texttimes}\mathrm{po}
\mbox{-}
\text{\texttimes}\mathrm{seq}$
that the sequence policy 
is able to produce reasonable sequence orders. 
We observe that PO is again 
able to improve performance.

After processing the packing results of the agent of
$\text{\texttimes}\mathrm{po}
\mbox{-}
\text{\texttimes}\mathrm{seq}
\mbox{-}
\text{\texttimes}\mathrm{att}
\mbox{-}
\text{\texttimes}\mathrm{leftover}$
with the aforementioned procedures,
we compare against the performance of
\cite{zhao2020online},
which packs on average 
$12.2$ items with a utility of
$54.7\%$
(with a lookahead number of $5$,
the utility achieved by their method
still does not surpass 
$60\%$);
while the strictly online-BPP version 
of our method on average
is able to pack
$15.6$ items with a utility of
$67.0\%$,
achieving state-of-the-art performance
in the online-BPP domain.
Visualizations of the packing results are shown in
Fig.\ref{fig:random-online-ablation-visual-b}-\ref{fig:random-online-after}.

\subsection{2D/3D BPP on Boxes Generated by Cut}
\label{sec:cut-rr}

As introduced in Sec.\ref{sec:cut-rr-ablation},
\cite{laterre2018ranked} (RR)
presented the state-of-the-art performance
on datasets generated by cut 
in both 
the $2\mathrm{D}$ and
the $3\mathrm{D}$ domain.
More specifically,
the boxes are generated 
by cutting a
$10\times10\times10$
bin
(or a 
$10\times10$
square
for $2\mathrm{D}$)
into 
$10$,
$20$,
$30$ or
$50$
boxes
(pseudo-code in Appendix \ref{app:cut-algo}).
They use a different reward than untility
$r_{\mathrm{u}}$
defined as
$r_{\mathrm{RR}}
=
\nicefrac{2\cdot(\sum_{n=1}^{N}l^n w^n)^{\nicefrac{1}{2}}}
{(L_{\pi} + W_{\pi})}$
for 
$2\mathrm{D}$
and
$r_{\mathrm{RR}}
=
\nicefrac{3\cdot(\sum_{n=1}^{N}l^n w^n h^n)^{\nicefrac{1}{3}}}
{(L_{\pi}W_{\pi} + W_{\pi}H_{\pi} + H_{\pi}L_{\pi})}$
for 
$3\mathrm{D}$.
For the following experiments,
we still train our method with the cost of
$1-r_{\mathrm{u}}$
but we compare the results in 
$r_{\mathrm{RR}}$.
We present the
visualizations of the packing results are shown in 
Fig.\ref{fig:cut-rr},
and the final testing results
in Table \ref{tab:cut-rr}.
We can observe that our method achieves
state-of-the-art in this data domain.

\subsection{3D BPP on Randomly Generated Boxes}
\label{sec:random-cql}

We continue to conduct 
comparisons
against the state-of-the-art algorithm 
\cite{jiang2021solving}
on randomly generated boxes,
in which
$W\times H=100\times100$
for the bin,
while the edges of boxes are sampled
from integers in the range of
$[20\twodots80]$.
\cite{jiang2021solving}
present experimental results
with box numbers of
$20$, 
$30$, 
$50$ 
and
$100$,
on which the learning curves of
our full method
$\mathrm{attend2pack}$
are shown in
Fig.\ref{fig:random-cql-visual-a}.
We can see that the performance improves stably during training, 
except for one of the trainings for 
$100$ boxes which encounters catastrophic forgetting.
We suspect that for more number of boxes
each episode would involve more rollout steps
thus an undesirable interaction between 
$\pi^{\mathrm{s}}$
and
$\pi^{\mathrm{p}}$
is more likely to propagate 
to unexpected gradient updates.
Since we observe larger variations
for the training with 
$100$ boxes,
here we add one more run to 
give a more accurate presentation 
of the performance of our algorithm on this task.
We present the visualizations of the packing results in
Fig.\ref{fig:random-cql-visual-b}-\ref{fig:random-cql-visual-d},
and the final testing results on 
$1\mathrm{e}{4}$
testing samples in
Table \ref{tab:random-cql},
where we can see that our method
achieves state-of-the-art performance 
in this data domain.

\subsection{Conclusions and Future Work}
\label{sec:conclusion}

To conclude,
this paper proposes a new model 
$\mathrm{attend2pack}$
that achieves state-of-the-art performance on BPP.
Possible future directions include 
extending this framework to 
multi-bin settings,
evaluating it under more packing constraints,
and more sophisticated solutions for credit assignment.


\clearpage




\bibliography{attend2pack}
\bibliographystyle{icml2021}

\clearpage
\appendix




\begin{figure}[t]
    \centering
	\includegraphics[width=0.99\columnwidth]{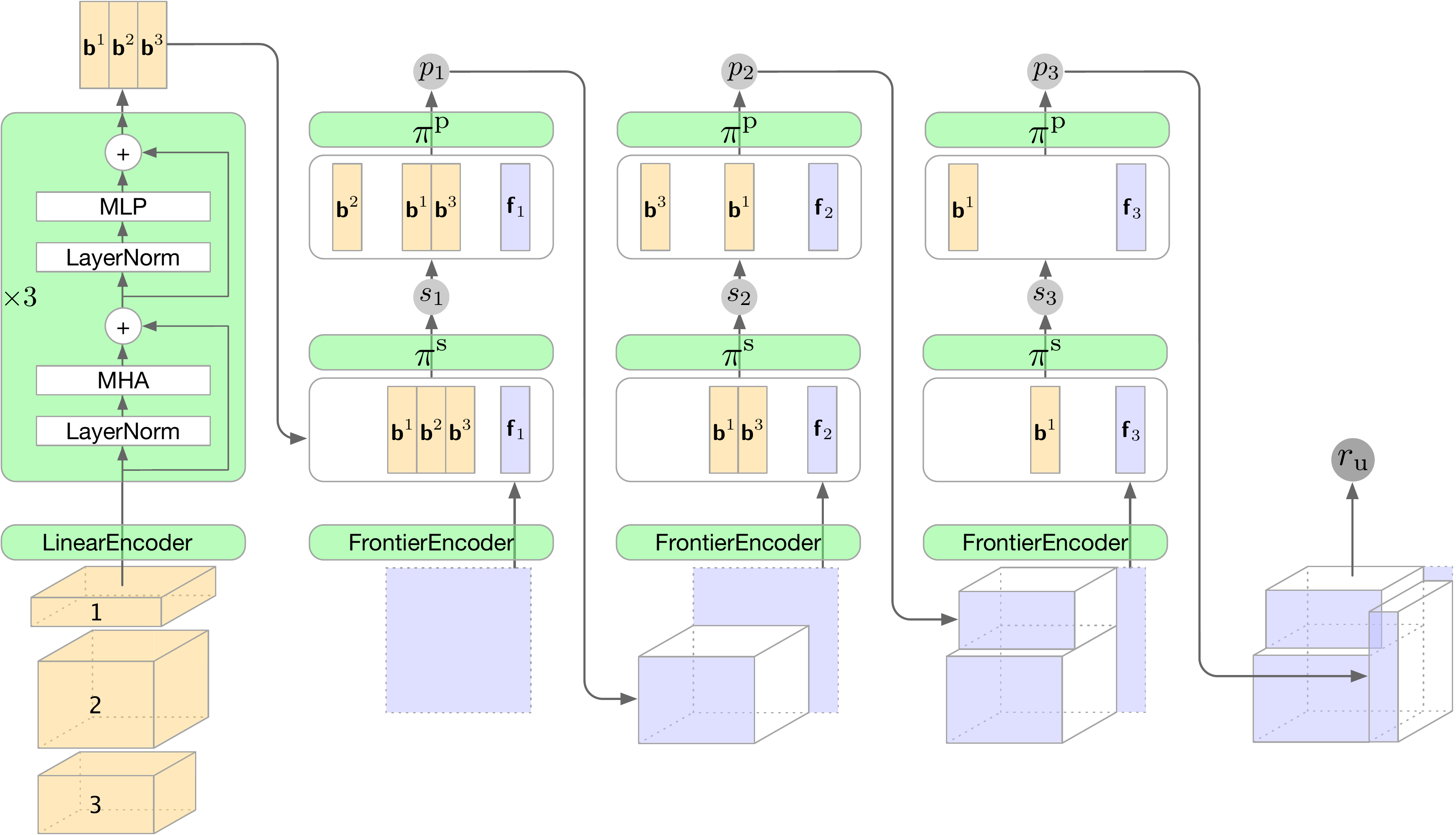}
    \caption{
Schematic plot
of the proposed 
$\mathrm{attend2pack}$
pipeline.
}
	\label{fig:pipeline}
\end{figure}

\section{Appendix: Schematic Plot of $\mathrm{attend2pack}$}
\label{app:pipeline}
A schematic plot
of our proposed $\mathrm{attend2pack}$
algorithm is shown in Fig.\ref{fig:pipeline}.

\section{Appendix: Schematic Plot of Glimpse}
\label{app:glimpse}
\begin{figure}[b]
    \centering
	\includegraphics[width=0.65\columnwidth]{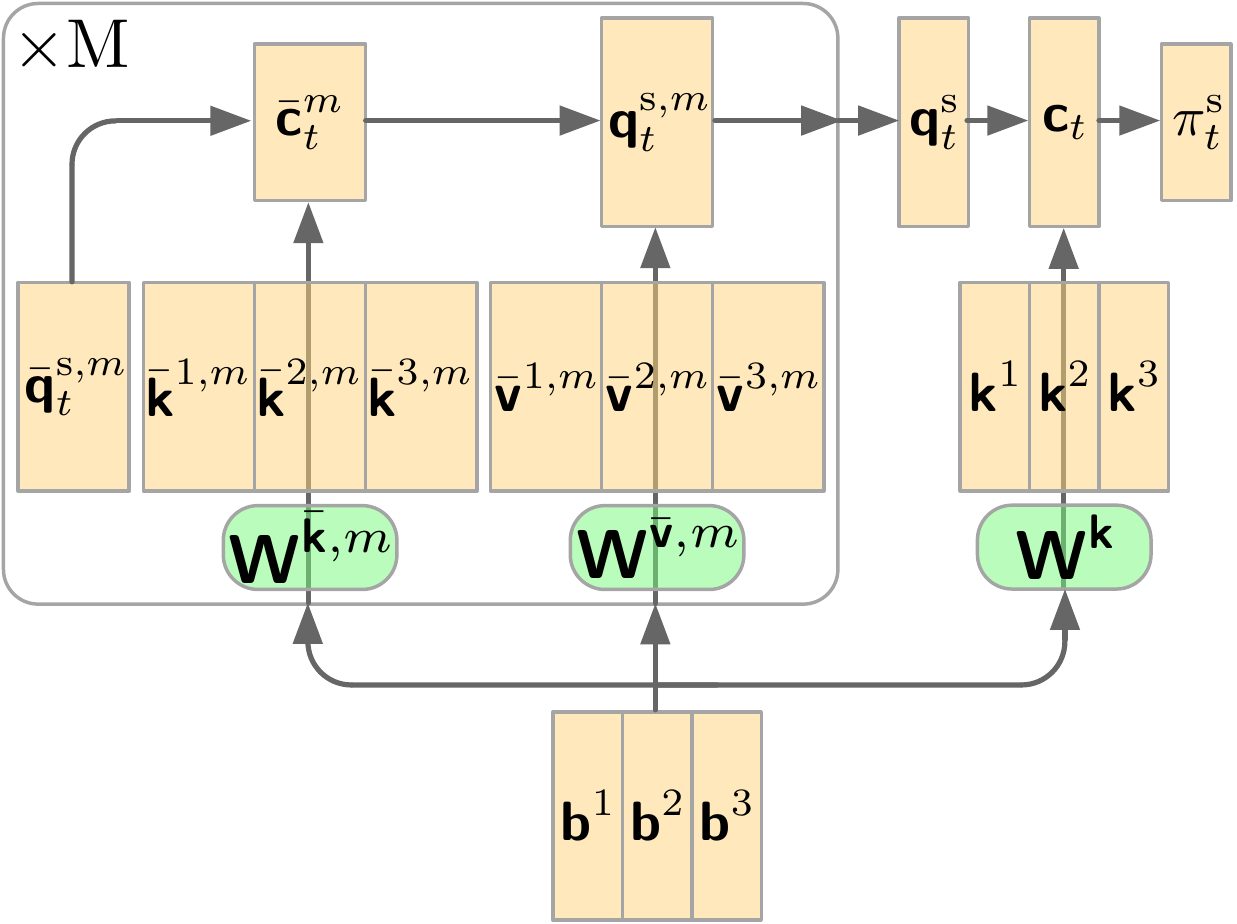}
    \caption{
Schematic plot
of the glimpse operation 
during sequence decoding
(Sec.\ref{sec:decoding-sequence}:
Eq.\ref{eq:g-proj}-\ref{eq:s-pi}).
}
	\label{fig:glimpse}
\end{figure}

A schematic plot
of the glimpse operation 
(Eq.\ref{eq:g-proj}-\ref{eq:s-pi})
during sequence decoding
is shown in Fig.\ref{fig:glimpse}.

\section{Appendix: Network Architecture}
\label{app:network}
In this section we present the details 
of the network architecture used in our proposed
$\mathrm{attend2pack}$ model.

The embedding network $\bm{\theta}^{\mathrm{e}}$ firstly encodes 
a $N\times3$ input into a $N\times128$ embedding
for each data point  
using an initial linear encoder with $128$ hidden units
(in preliminary experiments,
we found that applying random permutation to the input list of boxes
gave at least marginal performance gains
over sorting the boxes,
so we adhere to random permutation in all our experiments;
also we ensure that all boxes are rotated such that
$l^n\le w^n \le h^n$
before they are fed into the network).
This is then followed by a multi-head self-attention module that contains 
$3$ self-attention layers with $8$ heads,
with each layer containing a self-attention sub-layer (Eq.\ref{eq:att-att}) 
and an $\mathrm{MLP}$ sub-layer (Eq.\ref{eq:att-mlp}). 
The self-attention sub-layer contains 
a $\mathrm{LayerNorm}$ followed by 
a standard $\mathrm{MHA}$ module \cite{vaswani2017attention} 
with an embedding dimension of $128$ and $8$ heads,
whose output is added to the input of the self-attention sub-layer with a skip connection.
The $\mathrm{MLP}$ module contains a fully-connected layer with $512$ hidden nodes
followed by $\mathrm{ReLU}$ activation,
which is then followed by another fully-connected layer with $128$ hidden nodes.
The output of the last fully-connected layer is added 
to the input of the $\mathrm{MLP}$ sub-layer with a skip connection.

As for the frontier encoding network,
for a bin with $W\times H=10\times10$ in the $3\mathrm{D}$ situation
(Sec.\ref{sec:cut-rr-ablation},
\ref{sec:random-online-ablation},
\ref{sec:cut-rr}),
the stacked two lastest frontiers (Sec.\ref{sec:frontier-embedding})
of size $2\times10\times10$ is frstly fed into 
a $2\mathrm{D}$ convolutional layer 
with $2$ input channels 
and $8$ output channels,
with kernel size, stride and padding 
of $3$, $1$ and $1$ respectively.
This is followed by $\mathrm{LayerNorm}$ and $\mathrm{ReLU}$ activation.
Then there is another convolutional layer with 
number of input channels of $8$ 
and number of output channels of $8$,
and with kernel size, stride and padding 
of $5$, $1$ and $1$ respectively.
This is again followed by $\mathrm{LayerNorm}$ and $\mathrm{ReLU}$ activation.
The output is then flattened into a vector of size $512$,
which is then passed into a fully-connected layer with $128$
hidden units,
which is again followed by $\mathrm{LayerNorm}$ and $\mathrm{ReLU}$ activation.

For the $2\mathrm{D}$ situation with a $2\mathrm{D}$ bin with $W=10$
(Sec.\ref{sec:cut-rr}),
the stacked last two frontier is
of size $2\times10$,
and is firstly fed into a $1\mathrm{D}$ convolutional layer
with number of input channels $2$ and 
number of output channels $8$,
with kernel size, stride and padding 
of $3$, $1$ and $1$ respectively.
This is followed by $\mathrm{LayerNorm}$ and $\mathrm{ReLU}$ activation.
The output is then flattened into a vector of size $80$,
then passed into a fully-connected layer with $128$
hidden units,
which is again followed by $\mathrm{LayerNorm}$ and $\mathrm{ReLU}$ activation.

For a bin with $W\times H=100\times100$ in the $3\mathrm{D}$ situation
(Sec.\ref{sec:random-cql}),
the stacked last two frontier
of size $2\times100\times100$
is firstly fed into a $2\mathrm{D}$ convolutional layer
with number of input channels $2$ and 
number of output channels $4$,
with kernel size, stride and padding 
of $3$, $2$ and $1$ respectively.
Then there is another two convolutional layers both with 
number of input channels of $4$ 
and number of output channels of $4$,
and with kernel size, stride and padding 
of $3$, $2$ and $1$ respectively.
This is again followed by $\mathrm{LayerNorm}$ and $\mathrm{ReLU}$ activation.
The output is then flattened into a vector of size $676$,
which is then passed into a fully-connected layer with $128$
hidden units,
which is again followed by $\mathrm{LayerNorm}$ and $\mathrm{ReLU}$ activation.
We note that that in the experiments in Sec.\ref{sec:random-cql},
for the experiments with 
$100$ number of boxes with bin of $W\times H=100\times100$,
instead of using the stacked last $2$ frontiers to be the input 
to the frontier encoding network as in all our other experiments,
we only use the most recent frontier due to limited memory. 

The sequence decoding parameters 
$\bm{\theta}^{\mathrm{s}}$ 
are presented 
in details in Sec.\ref{sec:decoding-sequence}.
We note an implementation detail is omitted in 
Sec.\ref{sec:decoding-sequence} and \ref{sec:decoding-placement},
that for both the encoding functions  
$f_{\mathcal{I}^{\mathrm{s}}}$ (Eq.\ref{eq:fs}) and
$f_{\mathcal{I}^{\mathrm{p}}}$ (Eq.\ref{eq:fp}),
each of their input component 
will go through a linear projection
before passing through the final 
$\langle\rangle$ operation
\begin{align}
    &f^{\mathrm{s}}_{\mathcal{I}}(s_{1:t-1}, p_{1:t-1}; \bm{\theta}^{\mathrm{e}})
=
\notag\\
    &\hspace{30pt}
    \big\langle
        \bm{\mathsf{W}}^{\mathrm{s}\text{-}\mathrm{leftover}}
        \langle
            \mathcal{B}
                \setminus{
                    \bm{\mathsf{b}}^{s_{1:t-1}} 
                }
        \rangle,
        \bm{\mathsf{W}}^{\mathrm{frontier}}
        \bm{\mathsf{f}}_{t}
    \big\rangle,
\\
    &f^{\mathrm{p}}_{\mathcal{I}}(s_{1:t}, p_{1:t-1}; \bm{\theta}^{\mathrm{e}})
=
\notag\\
    &\hspace{30pt}
    \big\langle
        \bm{\mathsf{W}}^{\mathrm{selected}}
        \bm{\mathsf{b}}^{s_t}, 
        \bm{\mathsf{W}}^{\mathrm{p}\text{-}\mathrm{leftover}}
        \langle
            \mathcal{B}
            \setminus{
                \bm{\mathsf{b}}^{s_{1:t}} 
            }
        \rangle,
        \bm{\mathsf{W}}^{\mathrm{frontier}}
        \bm{\mathsf{f}}_{t}
    \big\rangle,
\end{align}
where all the weight matrices 
$\bm{\mathsf{W}}$
are of size 
$128\times128$.

As for the placement decoding network 
$\bm{\theta}^{\mathrm{p}}$,
it fistly contains a fully-connected layer with 
$128$ hidden units followed 
by $\mathrm{LayerNorm}$ and $\mathrm{ReLU}$ activation.
This is followed by another fully-connected layer with 
$128$ hidden units 
with $\mathrm{LayerNorm}$ and $\mathrm{ReLU}$ activation.
Then the final output fully-connected layer maps
$128$-dimensional embeddings to placement logits.
The logit vector is of size 
$6\times10$ for 
$3\mathrm{D}$ packing with bin of $W\times H=10\times10$
($6$ possible orientations)
(Sec.\ref{sec:cut-rr-ablation},
\ref{sec:random-online-ablation},
\ref{sec:cut-rr}),
$2\times10$ for 
$2\mathrm{D}$ packing with $W=10$
($2$ possible orientations)
(Sec.\ref{sec:cut-rr}),
and $6\times100$ for 
$3\mathrm{D}$ packing with bin of $W\times H=100\times100$
($6$ possible orientations)
(Sec.\ref{sec:random-cql}).

\begin{figure}[t]
    \centering
    \begin{subfigure}{0.87\columnwidth}
        \includegraphics[width=\columnwidth]{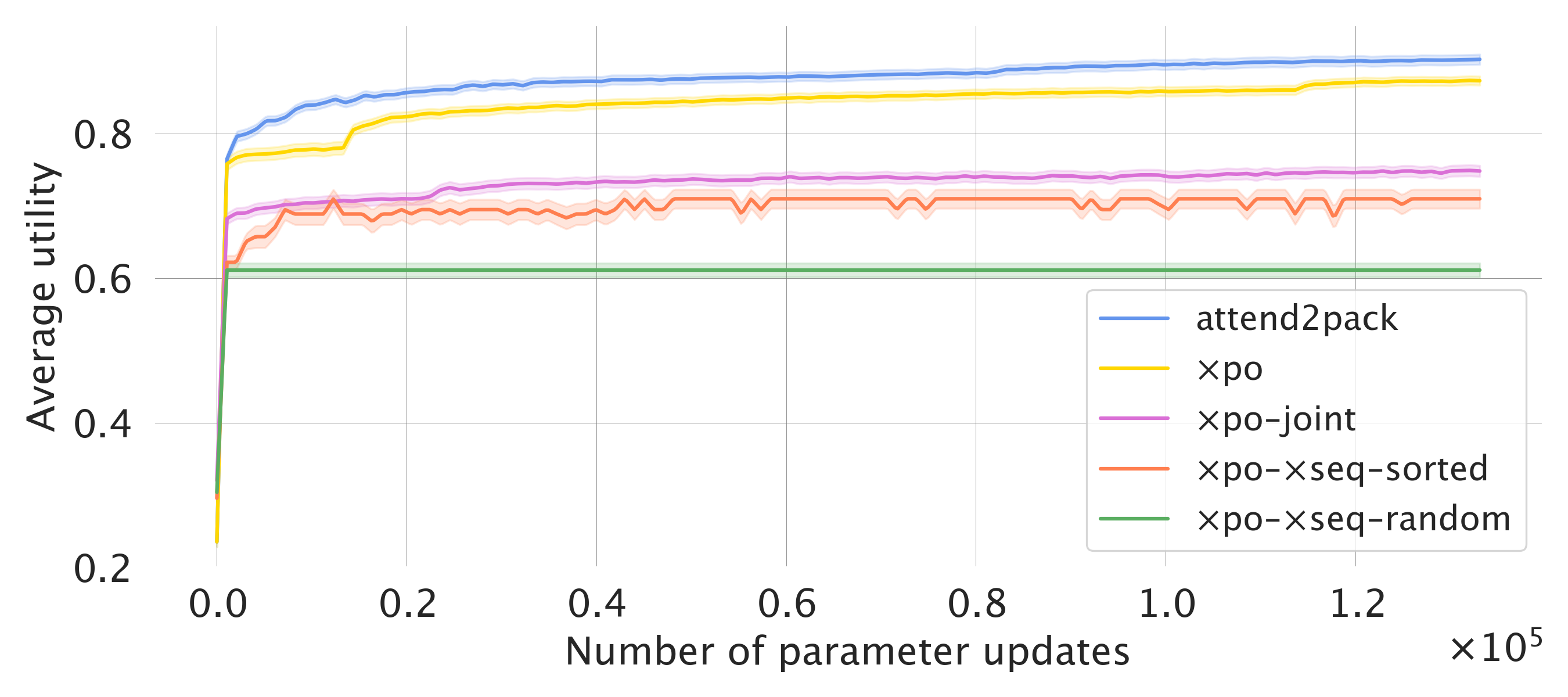}
        \caption{
Additional experiments on the dataset used in 
Sec.\ref{sec:cut-rr-ablation}.
        }
        \label{fig:cut-rr-seq-ablation}
    \end{subfigure}
    \begin{subfigure}{0.87\columnwidth}
        \includegraphics[width=\columnwidth]{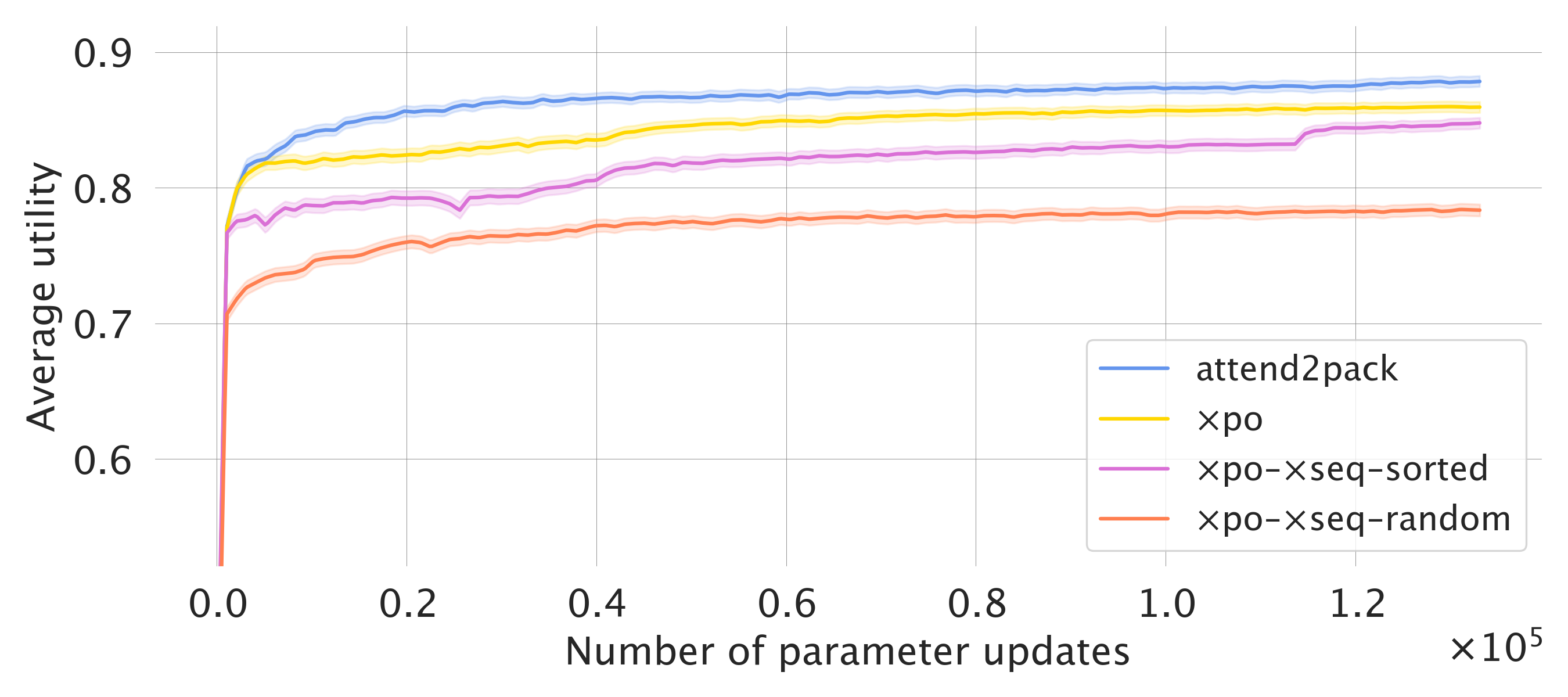}
        \caption{
Additional experiments on the dataset used in 
Sec.\ref{sec:random-online-ablation}
(we note that here
$\text{\texttimes}\mathrm{po}
\mbox{-}
\text{\texttimes}\mathrm{seq}
\mbox{-}
\mathrm{random}$
corresponds to
$\text{\texttimes}\mathrm{po}
\mbox{-}
\text{\texttimes}\mathrm{seq}$
in Fig.\ref{fig:random-online-ablation-visual}).
            }
        \label{fig:random-online-seq-ablation}
    \end{subfigure}
    \caption{
Results for App.\ref{app:seq-ablation},
where we plot the average utility obtained in evaluation during training,
each plot shows mean $\pm\hspace{3pt}0.1\hspace{2pt}\cdot$ standard deviation.
}
    \label{fig:seq-ablation}
\end{figure}

\section{Appendix: Additional Ablation Study on the Sequence Policy}
\label{app:seq-ablation}
We conduct an additional set of comparisons
on top of the ablation studies in 
Sec.\ref{sec:cut-rr-ablation},\ref{sec:random-online-ablation}.
Specifically,
we compare additionally with training setups
where the sequential order of the boxes are not learned,
but is fixed by random permutation or 
sorted by a simple heuristic:
sorted from large to small
by volumn $l^n\cdot w^n\cdot h^n$,
where ties are broken by
$\nicefrac{w^n}{h^n}$
then by
$\nicefrac{l^n}{w^n}$
(we note that 
$l^n\le w^n \le h^n$ (App.\ref{app:network})).
The setups without learning the sequence order
are denoted as
$\text{\texttimes}\mathrm{po}
\mbox{-}
\text{\texttimes}\mathrm{seq}
\mbox{-}
\mathrm{random}$
(fixed random order)
and
$\text{\texttimes}\mathrm{po}
\mbox{-}
\text{\texttimes}\mathrm{seq}
\mbox{-}
\mathrm{sorted}$
(fixed sorted order)
respectively.
The experimental results are shown in
Fig.\ref{fig:seq-ablation},
from which we can observe that
for the random dataset 
(Fig.\ref{fig:random-online-seq-ablation}),
the simple heuristic of following 
a sorted sequential order
could lead to satisfactory performance 
together with a learned placement policy;
while this sorted heuristic could not yield
strong performance for the cut dataset
(Fig.\ref{fig:cut-rr-seq-ablation}).
Therefore learning the sequence policy should be 
considered when dealing with an arbitrary dataset configuration.

\begin{figure}[t]
	\centering
	\begin{subfigure}{0.20\columnwidth}
		\centering
		\includegraphics[scale=0.15]{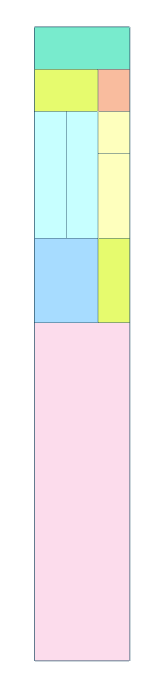}
		\caption{$L=3$}
	\end{subfigure}
	\begin{subfigure}{0.30\columnwidth}
		\centering
		\includegraphics[scale=0.15]{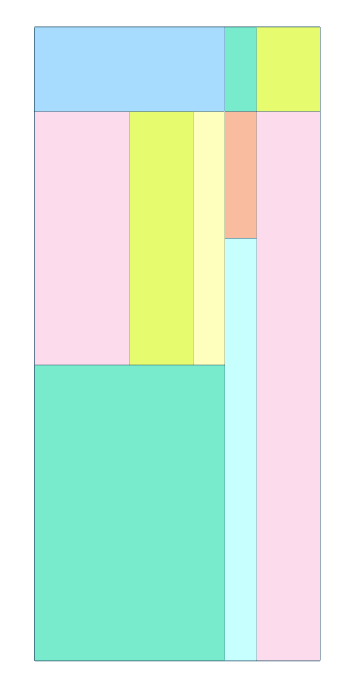}
		\caption{$L=9$}
	\end{subfigure}
	\begin{subfigure}{0.40\columnwidth}
		\centering
		\includegraphics[scale=0.15]{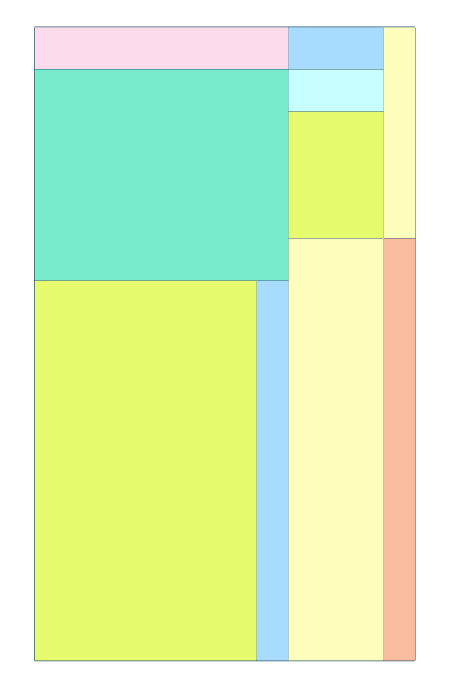}
		\caption{$L=12$}
	\end{subfigure}
	\caption{
Visualizations for App.\ref{app:cut-r2},
on dataset generated by cutting 
a rectangle
into
$10$
boxes,
with
$W=15$
and $L$
uniformly sampled from
$[2\twodots15]$,
where the box edges range within
$[1\twodots15]$.
	}
	\label{fig:cut-r2-visual}
\end{figure}
\begin{table}[b]
  \caption{
  Comparisons in 
  Sec.\ref{app:cut-r2}
  on $r_{L}$
  against
  $\mathrm{HVRAA}$\cite{zhu2017three},
  Lego heuristic\cite{hu2017solving},
  $\mathrm{MCTS}$\cite{browne2012survey} and
  $\mathrm{RR}$\cite{laterre2018ranked}.
  The 
  $r_{\mathrm{u}}$
  obtained by our method is presented in the last row.
  }
  \label{tab:cut-r2}
  \begin{center}
    \begin{scriptsize}
      \begin{sc}
        \begin{tabular}{ccccccc}
          \toprule
          HVRAA                          & $0.896\pm0.239$ \\
          Lego heuristic                 & $0.737\pm0.349$ \\
          MCTS                           & $0.936\pm0.097$ \\
          RR                             & $0.964\pm0.160$ \\
          attend2pack                    & $\bm{0.971}\pm0.051$ \\
          \midrule
          attend2pack ($r_{\mathrm{u}}$) & $\bm{0.971}\pm0.051$ \\
          \bottomrule
        \end{tabular}
      \end{sc}
    \end{scriptsize}
  \end{center}
\end{table}

\section{Appendix: Additional Experiments on 2D BPP on Boxes Generated by Cut}
\label{app:cut-r2}
We conduct an additional set of experiments comparing against
\cite{wang2021self},
which adopts the 
$\mathrm{RR}$
algorithm on a 2D BPP dataset generated by cutting
2D rectangles 
into $10$ boxes,
with 
$W=15$
and 
$L$ uniformly sampled from 
$[2\twodots15]$,
where the box edges are guaranteed to be within 
$[1\twodots15]$.
The reward measure used in 
\cite{wang2021self} is
$r_{L}=\nicefrac{L_{\pi}}
{\frac{(\sum_{n=1}^N l^n w^n)}{W}}$,
where
$L_{\pi}$
denotes the length of the minimum bounding box
that encompass all packed boxes at the end of episodes;
as in Sec.\ref{sec:cut-rr},
we train our agent
$\mathrm{attend2pack}$
with the utility reward 
$r_{\mathrm{u}}$
but compare to the results of 
\cite{wang2021self}
using their measure 
$r_{L}$.
The packing visualizations and the final testing results
are shown in 
Fig.\ref{fig:cut-r2-visual} and
Table \ref{tab:cut-r2}.

\begin{figure}[t]
	\centering
	\begin{subfigure}{0.87\columnwidth}
		\includegraphics[width=\columnwidth]{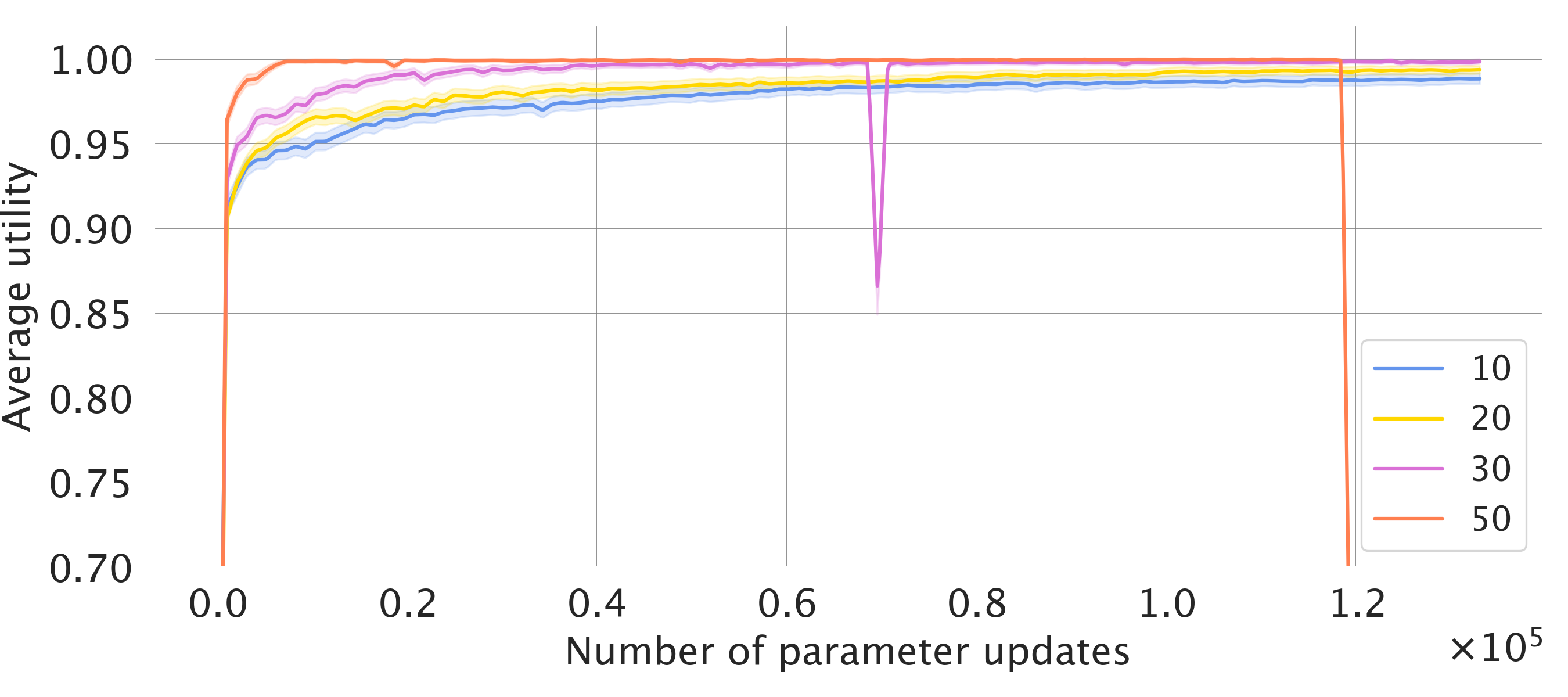}
		\caption{
Average utility obtained in evaluation during training,
each plot shows mean $\pm\hspace{3pt}0.1\hspace{2pt}\cdot$ standard deviation.
			}
	\end{subfigure}

	\begin{subfigure}{0.23\columnwidth}
		\includegraphics[width=\columnwidth]{figs/052800-cut-rr-2-10-waste-3.0-permute-f2-1e-4-po1-43-0-crop.png}
		\caption{$10$ items}
	\end{subfigure}
	\begin{subfigure}{0.23\columnwidth}
		\includegraphics[width=\columnwidth]{figs/052800-cut-rr-2-20-waste-3.0-permute-f2-1e-4-po1-42-0-crop.png}
		\caption{$20$ items}
	\end{subfigure}
	\begin{subfigure}{0.23\columnwidth}
		\includegraphics[width=\columnwidth]{figs/052800-cut-rr-2-30-waste-3.0-permute-f2-1e-4-po1-43-0-crop.png}
		\caption{$30$ items}
	\end{subfigure}
	\begin{subfigure}{0.23\columnwidth}
		\includegraphics[width=\columnwidth]{figs/052800-cut-rr-2-50-waste-3.0-permute-f2-1e-4-po1-43-0-crop.png}
		\caption{$50$ items}
	\end{subfigure}

	\begin{subfigure}{0.87\columnwidth}
		\includegraphics[width=\columnwidth]{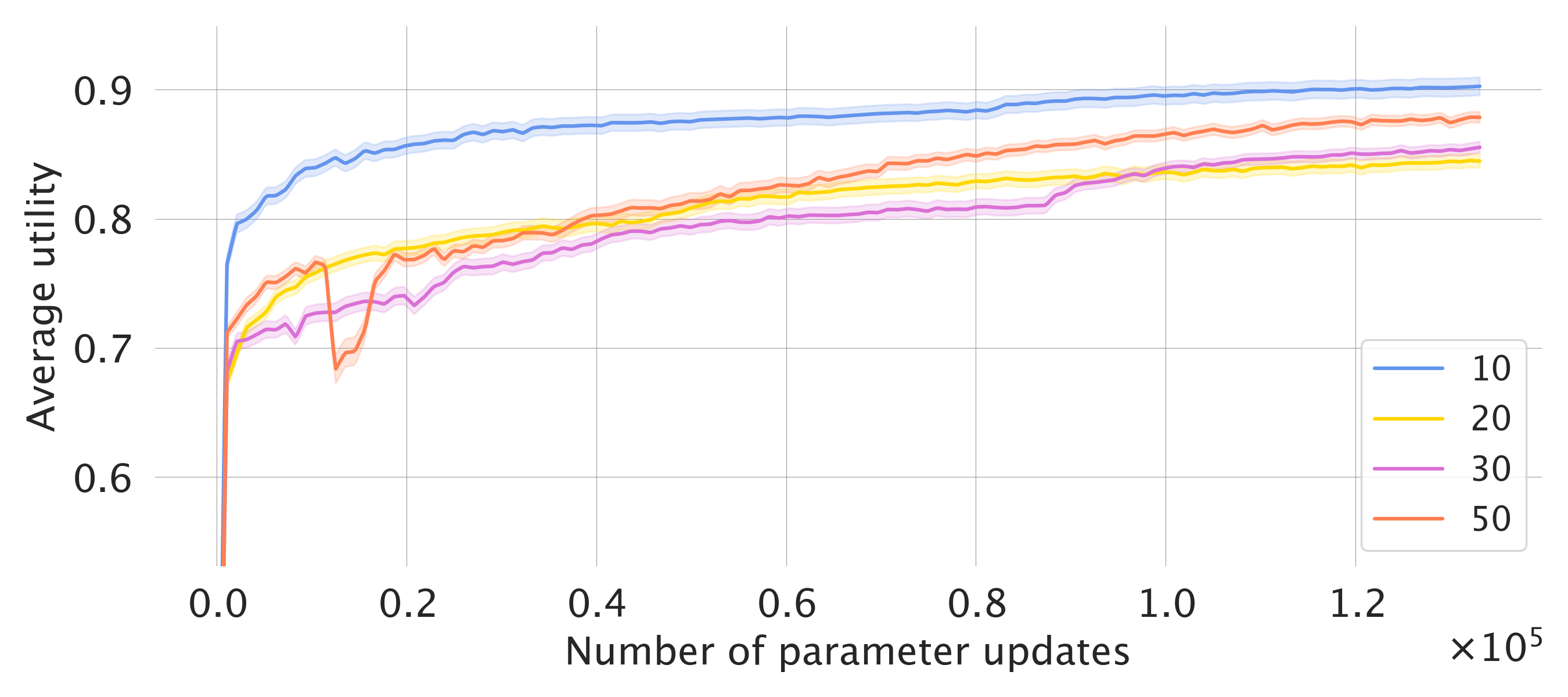}
		\caption{
Average utility obtained in evaluation during training,
each plot shows mean $\pm\hspace{3pt}0.1\hspace{2pt}\cdot$ standard deviation.
			}
	\end{subfigure}

	\begin{subfigure}{0.24\columnwidth}
		\includegraphics[width=\columnwidth]{figs/052800-cut-rr-3-10-waste-3.0-permute-f2-1e-4-po1-42-7-crop.png}
		\caption{$10$ items}
	\end{subfigure}
	\begin{subfigure}{0.24\columnwidth}
		\includegraphics[width=\columnwidth]{figs/052800-cut-rr-3-20-waste-3.0-permute-f2-1e-4-po1-43-16-crop.png}
		\caption{$20$ items}
	\end{subfigure}
	\begin{subfigure}{0.24\columnwidth}
		\includegraphics[width=\columnwidth]{figs/052800-cut-rr-3-30-waste-3.0-permute-f2-1e-4-po1-42-1-crop.png}
		\caption{$30$ items}
	\end{subfigure}
	\begin{subfigure}{0.24\columnwidth}
		\includegraphics[width=\columnwidth]{figs/052800-cut-rr-3-50-waste-3.0-permute-f2-1e-4-po1-42-1-crop.png}
		\caption{$50$ items}
	\end{subfigure}

	\caption{
Visualizations for Sec.\ref{sec:cut-rr},
on dataset generated by cutting 
a square of $10\times10$ ($2\mathrm{D}$) 
or a bin of $10\times10\times10$ ($3\mathrm{D}$)
into
$10$,
$20$,
$30$,
$50$
items,
with the box edges ranging within
$[1\twodots10]$.
	}
	\label{fig:cut-rr-app}
\end{figure}

\begin{figure}[t]
	\centering
	\begin{subfigure}{0.45\columnwidth}
		\includegraphics[width=\columnwidth]{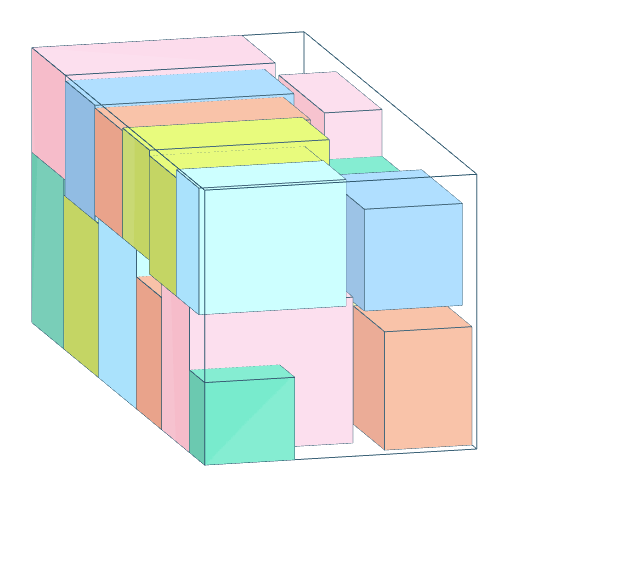}
		\caption{$20$ items}
	\end{subfigure}
	\begin{subfigure}{0.45\columnwidth}
		\includegraphics[width=\columnwidth]{figs/052800-random-cql-3-30-waste-3.0-permute-f2-1e-4-po1-43-6-crop.png}
		\caption{$30$ items}
	\end{subfigure}
	\begin{subfigure}{0.45\columnwidth}
		\includegraphics[width=\columnwidth]{figs/052800-random-cql-3-50-waste-3.0-permute-f2-1e-4-po1-43-9-crop.png}
		\caption{$50$ items}
	\end{subfigure}
	\begin{subfigure}{0.45\columnwidth}
		\includegraphics[width=\columnwidth]{figs/052800-random-cql-3-100-waste-3.0-permute-f1-1e-4-po1-43-18-crop.png}
		\caption{$100$ items}
	\end{subfigure}
	\caption{
Visualizations 
for Sec.\ref{sec:random-cql},
on dataset generated by randomly sampling
the edges of
$20$,
$30$,
$50$,
$100$
boxes
from
$[20\twodots80]$,
with a bin of 
$W\times H=100\times100$.
	}
	\label{fig:random-cql-visual-app}
\end{figure}
\section{Appendix: More Visualizations for Sec.\ref{sec:cut-rr}, \ref{sec:random-cql}}
Additional results for Sec.\ref{sec:cut-rr}, \ref{sec:random-cql} 
are shown in Fig.\ref{fig:cut-rr-app}, \ref{fig:random-cql-visual-app}.



\section{Appendix: Dataset Specifications}
We list the specifications of the datasets used 
in all our presented experiments in Table \ref{tab:specs}.
\begin{table}[b]
  \caption{
  Specifications of datasets 
  in our presented experiments.
  }
  \label{tab:specs}
  \begin{center}
    \begin{scriptsize}
      \begin{sc}
        \begin{tabular}{ccccc}
          \toprule
                & dataset type      & $L\times W(\times H)$    & $N$     & $l^i$,$w^i$,($h^i)$ \\
          \midrule
          Sec.\ref{sec:cut-rr-ablation}, App.\ref{app:seq-ablation}        & cut-3D    & $10\times10\times10$      & $10$           & $[1\twodots10]$  \\
          Sec.\ref{sec:random-online-ablation}, App.\ref{app:seq-ablation} & random-3D & $\cdot\times10\times10$   & $24$           & $[2\twodots5]$   \\
          Sec.\ref{sec:cut-rr}                 & cut-2D    & $10\times10$              & $10,20,30,50$  & $[1\twodots10]$  \\
          Sec.\ref{sec:cut-rr}                 & cut-3D    & $10\times10\times10$      & $10,20,30,50$  & $[1\twodots10]$  \\
          Sec.\ref{sec:random-cql}             & random-3D & $\cdot\times100\times100$ & $20,30,50,100$ & $[20\twodots80]$ \\
          App.\ref{app:cut-r2}                 & cut-2D    & $[2\twodots15]\times15$   & $10$           & $[1\twodots15]$  \\
          \bottomrule
        \end{tabular}
      \end{sc}
    \end{scriptsize}
  \end{center}
\end{table}

\section{Appendix: Inference Time}
We list the inference time 
of our presented experiments 
in Table \ref{tab:inference},
for which the CPU configuration of our computing resource is: 
Intel(R) Core(TM) i9-7940X CPU 3.10GHz 128GB,
and the GPU configuration is: 
Nvidia TITAN X (Pascal) GPU with 12G memory.

\begin{table}[t]
  \caption{
  Inference time for a batch of size $1024$
  on GPU 
  (Intel(R) Core(TM) i9-7940X CPU 3.10GHz 128GB with a Nvidia TITAN X (Pascal) GPU with 12G memory)
  of our presented experiments
  (we note that the some operations for 
  $\mathrm{3D}$ 
  are written in $\mathrm{cuda}$,
  while the rest of the code is in
  $\mathrm{python}$/$\mathrm{pytorch}$).
  }
  \label{tab:inference}
  \begin{center}
    \begin{scriptsize}
      \begin{sc}
        \begin{tabular}{c|c|c|c|c}
          \toprule
          \multicolumn{1}{l}{} & & $L\times W(\times H)$    & $N$ & Time(\si{\second}) \\
          \midrule
          2D & Sec.\ref{sec:cut-rr} & $10\times10$              & $10$ & 0.108\si{\second} \\
          2D & Sec.\ref{sec:cut-rr}                           & $10\times10$              & $20$ & 0.250\si{\second} \\
          2D & Sec.\ref{sec:cut-rr}                           & $10\times10$              & $30$ & 0.448\si{\second} \\
          2D & Sec.\ref{sec:cut-rr}                           & $10\times10$              & $50$ & 1.015\si{\second} \\
          2D & App.\ref{app:cut-r2}                           & $[2\twodots15]\times15$   & $10$ & 0.145\si{\second} \\
          \midrule
          3D & Sec.\ref{sec:random-online-ablation}, App.\ref{app:seq-ablation}            & $\cdot\times10\times10$   & $24$ & 0.224\si{\second} \\
          3D & Sec.\ref{sec:cut-rr-ablation},\ref{sec:cut-rr}, App.\ref{app:seq-ablation}  & $10\times10\times10$      & $10$ & 0.097\si{\second} \\
          3D & Sec.\ref{sec:cut-rr}                           & $10\times10\times10$      & $20$ & 0.185\si{\second} \\
          3D & Sec.\ref{sec:cut-rr}                           & $10\times10\times10$      & $30$ & 0.269\si{\second} \\
          3D & Sec.\ref{sec:cut-rr}                           & $10\times10\times10$      & $50$ & 0.504\si{\second} \\
          3D & Sec.\ref{sec:random-cql}                       & $\cdot\times100\times100$ & $20$ & 0.535\si{\second} \\
          3D & Sec.\ref{sec:random-cql}                       & $\cdot\times100\times100$ & $30$ & 0.805\si{\second} \\
          3D & Sec.\ref{sec:random-cql}                       & $\cdot\times100\times100$ & $50$ & 1.379\si{\second} \\
          3D & Sec.\ref{sec:random-cql}                       & $\cdot\times100\times100$ & $100$ & 2.821\si{\second} \\
          \bottomrule
        \end{tabular}
      \end{sc}
    \end{scriptsize}
  \end{center}
\end{table}

\section{Appendix: Cut Dataset Generation}
\label{app:cut-algo}

The algorithm from
\cite{laterre2018ranked} 
that cuts a fixed-sized bin into a designated number of cubic items 
is presented in Alg.\ref{alg:cut}.
This algorithm is used to generate the dataset used in the experiments in 
Sec.\ref{sec:cut-rr-ablation},
\ref{sec:cut-rr} and
App.\ref{app:cut-r2}.

\begin{algorithm}[t]
   \caption{Cutting Bin from \cite{laterre2018ranked}}
   \label{alg:cut-box}
   \begin{algorithmic}
      \STATE {\bfseries Input:} Bin dimensions: length $L$, width $W$, height $H$;
      minimum length for box edges $M$; number of boxes $N$ to cut the bin into.
      \STATE {\bfseries Output:} A set of $N$ boxes $\mathcal{S}$.
      \STATE Initialize box list $\mathcal{S} = \{(L, W, H)\}$.
      \WHILE{$\lvert \mathcal{S} \rvert < N$}
      \STATE {\bfseries Sample} a box $i$ $(l^i, w^i, h^i)$ from $\mathcal{S}$ 
         by probabilities proportional to the volumes of the boxes; 
         pop box $i$ from $\mathcal{S}$.
      \STATE {\bfseries Sample} a dimension (e.g. $w^i$) from $(l^i, w^i, h^i)$ 
         by probabilities proportional to the lengths of the three dimensions. 
      \STATE {\bfseries Sample} a cutting point $j$ from the selected dimension $w^i$ 
         by probabilities proportional to the distances from all points on $w^i$
         to the center of $w^i$.
         Check that $j\ge M$ and $w^i-j\ge M$,
         otherwise push the selected box $(l^i,w^i,h^i)$ back into $\mathcal{S}$ 
         and continue to the next iteration.
      \STATE {\bfseries Cut} box $(l^i, w^i, h^i)$ along $w^i$ from the chosen point $j$ 
         to get two boxes
         $(l^i, j, h^i)$ and $(l^i, w^i-j, h^i)$
         and push those two boxes into $\mathcal{S}$.
      \ENDWHILE
      \STATE return {\bfseries $\mathcal{S}$}
   \end{algorithmic}
   \label{alg:cut}
\end{algorithm}


\end{document}